\documentclass[11pt]{article}

\usepackage{authblk}

\usepackage{amsmath, amssymb, amsthm}
\usepackage{tikz}
\usepackage{graphicx}
\usepackage{hyperref}
\usepackage{bm}
\usepackage{geometry}
\usepackage{tikz}
\usepackage{booktabs}
\usepackage{subcaption,enumitem}
\usetikzlibrary{positioning, arrows.meta, fit}

\theoremstyle{plain}
\newtheorem{remark}{Remark}
\newtheorem{proposition}{Proposition}

\title{Parametric Numerical Integration with (Differential) Machine Learning}
\author[1,2,*]{Álvaro Leitao}
\author[2,1]{Jonatan Ráfales}

\affil[1]{CITIC Research center, Spain}
\affil[2]{Department of Mathematics, University of A Coruña, Spain}
\affil[*]{Corresponding author: alvaro.leitao@udc.gal}
\date{\today}

\begin{document}

\maketitle

\begin{abstract}
In this work, we introduce a machine/deep learning methodology to solve parametric integrals. Besides classical machine learning approaches, we consider a differential learning framework that incorporates derivative information during training, emphasizing its advantageous properties. Our study covers three representative problem classes: statistical functionals (including moments and cumulative distribution functions), approximation of functions via Chebyshev expansions, and integrals arising directly from differential equations. These examples range from smooth closed-form benchmarks to challenging numerical integrals. Across all cases, the differential machine learning-based approach consistently outperforms standard architectures, achieving lower mean squared error, enhanced scalability, and improved sample efficiency.
\end{abstract}

\section{Introduction}

Parametric integration arises when integral quantities depend on model (problem-wise) parameters, a common situation across statistics \cite{cramer1999, lehmann1998, rohatgi2015}, probability \cite{alon2016, durrett2019, feller1991}, approximation theory \cite{Szego1967, trefethen2019, zygmund2002}, and differential equations \cite{chandrasekhar2013, cont2003, ince1956}. Examples include, among others, moments, cumulative distribution functions (CDFs), and orthogonal polynomials' representation. These problems also appear in uncertainty quantification, Bayesian inference, and stochastic modelling, where integrals must be computed repeatedly for different parameter configurations \cite{ghanem2017, rafales2021, rafales2024, sullivan2015}. While some integrals admit closed-form solutions, many require evaluation of non-elementary expressions across wide parameter ranges.

Classical quadrature methods, such as Gaussian and Newton--Cotes rules, provide accurate solutions for low-dimensional problems \cite{davis2007, quarteroni2006}, but their cost grows rapidly with dimension or accuracy requirements. Recent advances in adaptive quadrature \cite{gander2000} and quasi-Monte Carlo quadrature \cite{kuo2011} extend the reach of these methods, but they remain limited for repeated high-dimensional parametric tasks.

Monte Carlo (MC) methods offer a natural alternative, approximating integrals through simulation averages and handling high-dimensional or irregular integrals more efficiently \cite{caflisch1998, owen2013}. They underpin modern Bayesian computation \cite{robert1999}, probabilistic numerics \cite{hennig2015}, and rare-event simulation \cite{glasserman2004}. However, evaluating the same integral across many parameter configurations can still be prohibitively expensive, especially in high-accuracy regimes. This motivates the search for surrogate models that approximate the mapping from parameters to integrals at negligible marginal cost once fitted.

Artificial Neural networks (ANNs) provide a flexible framework for such surrogates. By training on a given information of the problem over a range of parameters, they are able to learn the functional dependence directly \cite{cybenko1989, hornik1989}. Once trained, they decouple the expensive (offline) fitting procedure from the (online) evaluation, enabling rapid predictions for unseen parameter values. Recent works have demonstrated the success of deep ANNs as surrogates in scientific computing, uncertainty quantification, and stochastic modelling \cite{beck2019, han2018, liu2021, raissi2019, villarino2023}. 

Our approach builds on the key insight of \cite{huge2020}, originally developed in the context of financial derivative pricing and recently extended to a range of other financial and stochastic applications \cite{benth2024, detering2025}. The central idea is to exploit the equivalence between conditional expectations and regression estimation. Within this simulation-based framework, ANNs are trained on noisy MC realizations rather than fully converged integral values. Crucially, an entire training dataset can be produced at the cost of a single MC run per parameter configuration, since each realization already provides an unbiased estimate of the integral. This reformulation casts parametric integration as a regression problem, a setting extensively analysed in statistical learning theory \cite{shalev2014, vapnik1998}. 

Despite these advantages, standard ANN training often remains sample-inefficient and can be unstable or inaccurate when approximating oscillatory integrals, integrals involving special functions, or when high accuracy is required near boundaries \cite{elbrachter2021, horvath2021, jiang2024, mishra2023}. In order to overcome these limitations, we also consider a differential machine learning (DML) approach \cite{frandsen2022, huge2020}. By jointly fitting quantities of interest and their gradients, the ANNs inherit smoothness and regularity from the underlying mathematics. This reduces variance, enhances generalization, and accelerates convergence relative to conventional architectures. Recent developments demonstrate that DML can dramatically improve accuracy and efficiency in high-dimensional learning problems across finance, statistics, and stochastic control \cite{beck2021, buehler2019, detering2025}. Consequently, in this work, ANNs are trained not only on single-realization MC estimations of the integrand but also on their differentials with respect to model parameters.

Beyond generic integral evaluations, we focus on three classes of parametric quantities in our numerical experiments. First, we study parametric moments and CDFs, which provide concise characterizations of probability distributions across varying parameters. Second, we examine function approximation via Chebyshev expansions, whose coefficients yield highly accurate representations for smooth and oscillatory parametric functions. In this context, we evaluate the ability of both standard ANNs and DML models to learn multi-output mappings corresponding to these coefficients. These techniques have been of practical relevance in computational finance, including the valuation of Bermudan swaptions~\cite{gomez2025} and the computation of dynamic initial margin in counterparty credit risk modelling~\cite{villarino2026}. Finally, we consider parametric integrals that arise naturally from ordinary differential equations (ODEs) and partial integro-differential equations (PIDEs). 

The paper is organized as follows. Section~\ref{sec:preliminaries} introduces the problem setting and reviews ANN and DML frameworks. Section~\ref{sec:methodology} details the proposed methodology. Section~\ref{sec:results} presents numerical results on parametric moments, CDFs, Chebyshev expansions, and parametric integrals derived directly from differential equations. Section~\ref{sec:conclusion} concludes with a discussion of the key findings and implications.

\section{Preliminaries} \label{sec:preliminaries}

This section establishes the mathematical framework and notation for parametric numerical integration. We begin by formulating parametric integration problems and discussing classical numerical schemes. We then reinterpret MC estimation as a regression problem, which naturally motivates the use of ANNs as surrogate models. Finally, we introduce DML as an extension of this framework, enhancing standard ANNs through the inclusion of derivative information.

\subsection{Parametric Numerical Integration}

As starting point, we formalize the parametric integration problem. Let $(\mathcal{X},\mathcal{B},\nu)$ be a probability space, and let 
$f:\mathcal{X}\times\Theta\to\mathbb{R}$ be a measurable function depending on a parameter $\boldsymbol{\theta}\in\Theta\subseteq\mathbb{R}^q$. For each fixed $\boldsymbol{\theta}$, the corresponding parametric integral is defined as
\begin{equation}\label{eq:param_integral}
I(\boldsymbol{\theta}) = \int_{\mathcal{X}} f(x;\boldsymbol{\theta})\,\nu(\mathrm{d}x) 
= \mathbb{E}_{\nu}\big[f(X;\boldsymbol{\theta})\big],
\end{equation}
where $X$ is a random variable distributed according to~$\nu$. This formulation highlights the equivalence between integration and expectation. Hence, the computation of parametric integrals can be naturally interpreted as a regression problem, since each integral represents the conditional expectation of the integrand with respect to the measure $\nu$, see \cite{longstaff2001}. 

In general, such integrals cannot be evaluated analytically, particularly when the integrand is nonlinear, high-dimensional, or costly to evaluate. Consequently, one must rely on numerical approximation schemes.

Classical approaches include quadrature methods and MC simulation. Quadrature rules achieve high accuracy in low-dimensional settings but deteriorate rapidly with increasing dimension. MC-based methods, in contrast, exhibit a convergence rate of order~$N^{-1/2}$ that is independent of the dimension, where $N$ denotes the number of simulations employed in an estimator $\widehat{I}_N(\boldsymbol{\theta})\approx I(\boldsymbol{\theta})$ of the form
\begin{equation}\label{eq:numeric_param_integral}
\widehat{I}_N(\boldsymbol{\theta}) 
= \frac{1}{N}\sum_{n=1}^N f(X_n;\boldsymbol{\theta}), 
\qquad X_n\overset{\text{i.i.d.}}{\sim}\nu.
\end{equation}

When the integration domain is bounded and $\nu$ corresponds to the uniform measure on $\mathcal{X} = [a,b]$, the integral in~\eqref{eq:param_integral} can be expressed as
\begin{equation} \label{eq:boun_param_integral}
I(\boldsymbol{\theta}) 
= \frac{1}{b-a} \int_a^b f(x;\boldsymbol{\theta})\,\mathrm{d}x,
\end{equation}
and the corresponding MC approximation takes the familiar form
\[
\widehat{I}_N(\boldsymbol{\theta})
= \frac{b-a}{N} \sum_{n=1}^N f(U_n;\boldsymbol{\theta}),
\qquad U_n \sim \mathcal{U}([a,b]),
\]
which provides an unbiased estimator of \eqref{eq:boun_param_integral}.

\subsection{Artificial Neural Networks and Differential Machine Learning} \label{subsec:ann_dml}

In the parametric setting, repeatedly evaluating~\eqref{eq:numeric_param_integral} for multiple parameter values $\boldsymbol{\theta}$ can become computationally prohibitive. This motivates the construction of surrogate models that efficiently approximate the entire mapping 
$\boldsymbol{\theta} \mapsto I(\boldsymbol{\theta})$. 

An ANN surrogate provides a flexible parametric representation of this mapping. Let $\widehat{I}(\boldsymbol{\theta};\mathbf{w})$ denote an ANN with weights $\mathbf{w} \in \mathbb{R}^{\mathrm{m}}$, trained to approximate the ground-truth integral for all $\boldsymbol{\theta} \in \Theta$. In this work, we additionally employ a single-realization MC estimator ($N=1$), which remains unbiased while substantially reducing computational cost at the expense of increased variance. Under this formulation, training the ANN becomes a classical supervised learning problem: the model learns to map inputs $\boldsymbol{\theta}^{(j)}$ to noisy targets $\hat{y}^{(j)}$, where each label corresponds to a single-realization MC run serving as an unbiased estimator of the true integral.

Given a set of parameter–sample pairs $(x^{(j)};\boldsymbol{\theta}^{(j)})$ with $x^{(j)} \stackrel{\text{i.i.d.}}\sim \nu$, each sample provides a noisy label of the integral, 
\begin{equation*}
\hat{y}^{(j)} = f(x^{(j)};\boldsymbol{\theta}^{(j)}) = I(\boldsymbol{\theta}^{(j)}) + \varepsilon_j,
\end{equation*}
where $\varepsilon_j$ is a zero-mean noise term with variance 
$\operatorname{Var}_{\nu}[\varepsilon_j] = \operatorname{Var}_{\nu}[f(x^{(j)};\boldsymbol{\theta}^{(j)})]$. These noisy labels naturally induce a supervised learning dataset (of size $J$) such that
\[
\mathcal{D} = \big\{ (\boldsymbol{\theta}^{(j)}, \hat{y}^{(j)}) \big\}_{j=1}^{J},
\]
where the task is to learn a mapping from parameters to integral values.

Proposition~\ref{prop:dml_variance_reduction} establishes the theoretical basis for employing both function evaluations and parameter-based gradients as unbiased training data in ANN surrogates. The proof of this proposition is provided in Appendix~\ref{app:proof_prop}.

\begin{proposition}[Unbiasedness and variance reduction via differential training]
\label{prop:dml_variance_reduction}

Let $(\mathcal{X},\mathcal{B},\nu)$ be a probability space, and 
$f:\mathcal{X}\times\Theta \to \mathbb{R}$ a measurable function, with $\Theta\subset \mathbb{R}^q$ open. For $\boldsymbol{\theta}\in\Theta$, define
\[
I(\boldsymbol{\theta}) := \int_{\mathcal{X}} f(x;\boldsymbol{\theta}) \, \nu(\mathrm{d}x).
\]

Fix $\boldsymbol{\theta}\in\Theta$ and let $\Omega_{\boldsymbol{\theta}}\subset\Theta$ be a neighbourhood of $\boldsymbol{\theta}$.  
Assume:

\begin{enumerate}[label=(A\arabic*)]
  \item $f(\cdot;\boldsymbol{\theta})\in L^1(\nu)$, and there exists $g\in L^1(\nu)$ such that
  \[
  |f(x;\boldsymbol{\theta}')|\le g(x)
  \quad\forall\, \boldsymbol{\theta}'\in\Omega_{\boldsymbol{\theta}},~
  \forall\,x\in\mathcal{X}.
  \]

  \item $f(x;\cdot)\in C^1(\Omega_{\boldsymbol{\theta}})$ for $\nu$-a.e.\ $x\in\mathcal{X}$.

  \item There exists $h\in L^1(\nu)$ such that
  \[
  \|\nabla_{\!\boldsymbol{\theta}} f(x;\boldsymbol{\theta}')\|
  \le h(x)
  \quad\forall\,\boldsymbol{\theta}'\in\Omega_{\boldsymbol{\theta}},~
  \nu\text{-a.e.\ } x\in\mathcal{X}.
  \]
\end{enumerate}

Let $\{\boldsymbol{\theta}^{(j)}\}_{j=1}^J\subset\Omega_{\boldsymbol{\theta}}$ be training points, and let $x^{(1)},\dots,x^{(J)}\stackrel{\text{i.i.d.}}{\sim}\nu$. Define the single-realization MC labels and their input gradients,
\[
\hat{y}^{(j)} := f(x^{(j)};\boldsymbol{\theta}^{(j)}), \qquad \hat{g}^{(j)} := \nabla_{\!\boldsymbol{\theta}} \hat{y}^{(j)} = \nabla_{\!\boldsymbol{\theta}} f(x^{(j)};\boldsymbol{\theta}^{(j)}), \qquad j = 1,\dots,J.
\]

Let $\widehat{I}^{(\vartheta)}(\cdot;\mathbf{w})$ be a DML model estimator with $\widehat{I}^{(\vartheta)}(\cdot;\mathbf{w})\in C^1(\Omega_{\boldsymbol{\theta}})$ for each $\mathbf{w} \in \mathbb{R}^\mathrm{m}$. For $\vartheta\in[0,1]$, define the loss function
\begin{equation} \label{eq:combined_loss}
\mathcal{L}_\vartheta(\mathbf{w}) := \vartheta\,\mathcal{L}^{(\vartheta)}_{\mathrm{val}}(\mathbf{w})
  + (1-\vartheta)\,\mathcal{L}^{(\vartheta)}_{\mathrm{diff}}(\mathbf{w}).
\end{equation}
where 
\begin{align*}
\mathcal{L}^{(\vartheta)}_{\mathrm{val}}(\mathbf{w})
&= \frac{1}{J}\sum_{j=1}^J 
   \mathbb{E}_{\nu}\!\left[
     \big(\widehat{I}^{(\vartheta)}(\boldsymbol{\theta}^{(j)};\mathbf{w}) 
     - \hat{y}^{(j)}\big)^2
   \right],\\
\mathcal{L}^{(\vartheta)}_{\mathrm{diff}}(\mathbf{w})
&= \frac{1}{J}\sum_{j=1}^J 
   \mathbb{E}_{\nu}\!\left[
     \big\|\nabla_{\!\boldsymbol{\theta}}
       \widehat{I}^{(\vartheta)}(\boldsymbol{\theta}^{(j)};\mathbf{w})
     - \hat{g}^{(j)}\big\|^2
   \right].
\end{align*}

Then,

\begin{enumerate}[label=(\roman*)]

\item (Unbiasedness).
The estimator $\widehat{I}^{(\vartheta)}$ is unbiased:
\[
\mathbb{E}_{\nu}
\big[\,\widehat{I}^{(\vartheta)}(\boldsymbol{\theta}^{(j)})\,\big]
= I(\boldsymbol{\theta}^{(j)}),
\qquad
\mathbb{E}_{\nu}
\big[\,\nabla_{\!\boldsymbol{\theta}}
       \widehat{I}^{(\vartheta)}(\boldsymbol{\theta}^{(j)})\,\big]
= \nabla_{\!\boldsymbol{\theta}} I(\boldsymbol{\theta}^{(j)}), \qquad j=1,\dots,J.
\]

\item (Variance reduction). The differential learning reduces the averaged variance:
\[
\frac{1}{J}\sum_{j=1}^J \operatorname{Var}_{\nu}\!
\left(\widehat{I}^{(\vartheta)}(\boldsymbol{\theta}^{(j)})\right)
\;\le\;
\frac{1}{J}\sum_{j=1}^J \operatorname{Var}_{\nu}\!
\left(\widehat{I}^{(1)}(\boldsymbol{\theta}^{(j)})\right).
\]

\end{enumerate}
\end{proposition}

Proposition~\ref{prop:dml_variance_reduction} therefore provides the theoretical justification for using single-realization MC labels, together with input gradients, as supervised learning data for surrogate modelling.

Specifically:
\begin{enumerate}[label=(\alph*)]
    \item Each $\widehat{I}^{(\vartheta)}(\boldsymbol{\theta}^{(j)};\mathbf{w})$ is an unbiased estimator of $I(\boldsymbol{\theta}^{(j)})$, justifying the use of single-realization MC estimations ($f(x^{(j)};\boldsymbol{\theta}^{(j)})$) and their gradients ($\nabla_{\!\boldsymbol{\theta}} f(x^{(j)};\boldsymbol{\theta}^{(j)})$) as training dataset.
    \item Incorporating differential information in the loss function \eqref{eq:combined_loss} reduces the averaged variance of the DML estimator $\widehat{I}^{(\vartheta)}(\boldsymbol{\theta}^{(j)};\mathbf{w})$ respect to the basic ANN estimator $\widehat{I}^{(1)}(\boldsymbol{\theta}^{(j)};\mathbf{w})$.
\end{enumerate}

In this work, we set \(\vartheta = \dfrac{1}{1+\omega q}\) following \cite{gomez2025}, where $q$ is the parameter dimension and $\omega\geq0$ is a tuneable weight. Note that when $\omega = 0$, we recover the classical training, yielding results equivalent to those of a standard ANN. In practice, we set $\omega = 1/q$ for the DML framework, ensuring that both the value and the differential loss contribute equally to the total loss.

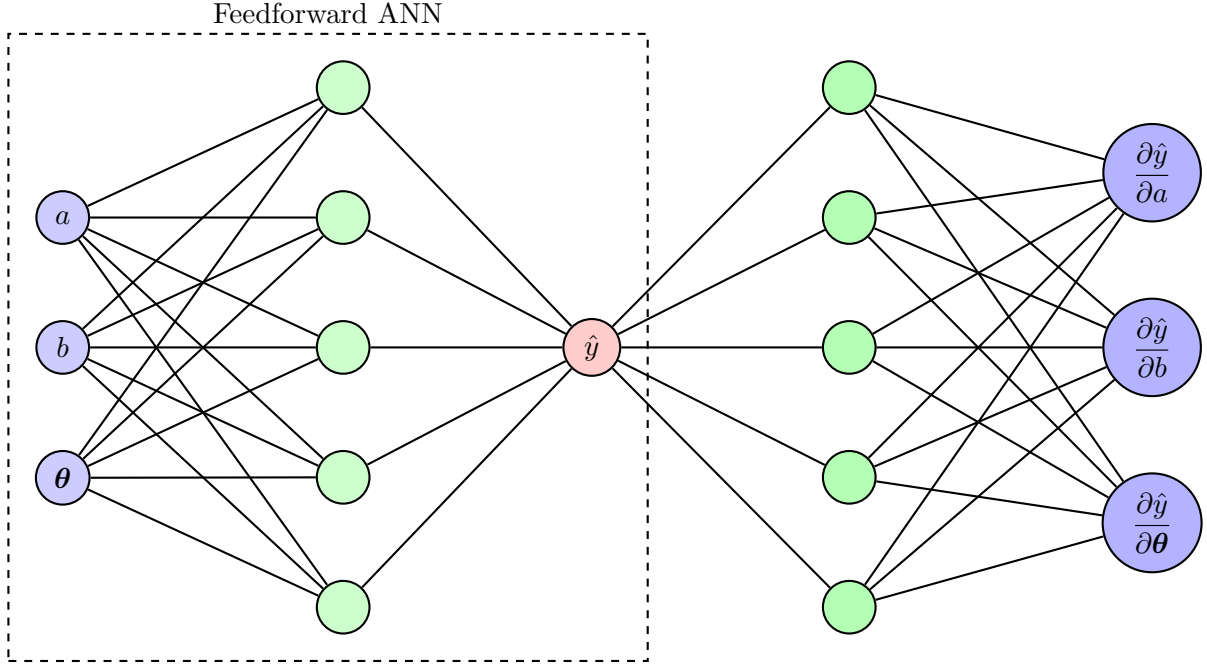
\begin{figure}[t!] 
\centering
\begin{tikzpicture}[
    node distance=1cm and 1cm,
    neuron/.style={circle, draw=black, thick, minimum size=7mm},
    line/.style={thick}
]

\node[neuron, fill=green!20] (H11) {};
\node[neuron, fill=green!20, below=of H11] (H12) {};
\node[neuron, fill=green!20, below=of H12] (H13) {};
\node[neuron, fill=green!20, below=of H13] (H14) {};
\node[neuron, fill=green!20, below=of H14] (H15) {};

\node[neuron, fill=blue!20, left=3cm of H12] (I1) {$a$};
\node[neuron, fill=blue!20, below=of I1] (I2) {$b$};
\node[neuron, fill=blue!20, below=of I2] (I3) {$\boldsymbol{\theta}$};

\node[neuron, fill=red!20, right=2.55cm of H13] (O1) {$\hat{y}$};

\node[neuron, fill=green!30, right=6cm of H11] (H21) {};
\node[neuron, fill=green!30, below=of H21] (H22) {};
\node[neuron, fill=green!30, below=of H22] (H23) {};
\node[neuron, fill=green!30, below=of H23] (H24) {};
\node[neuron, fill=green!30, below=of H24] (H25) {};

\node[neuron, fill=blue!30, right=3cm of H23] (O2) {$\dfrac{\partial\hat{y}}{\partial b}$};
\node[neuron, fill=blue!30, above=of O2] (O3) {$\dfrac{\partial\hat{y}}{\partial a}$};
\node[neuron, fill=blue!30, below=of O2] (O4) {$\dfrac{\partial\hat{y}}{\partial\boldsymbol{\theta}}$};

\foreach \i in {I1,I2,I3}
  \foreach \j in {H11,H12,H13,H14,H15}
    \draw[line] (\i) -- (\j);

\foreach \j in {H11,H12,H13,H14,H15}
  \draw[line] (\j) -- (O1);

\foreach \j in {H21,H22,H23,H24,H25}
  \draw[line] (O1) -- (\j);

\foreach \i in {H21,H22,H23,H24,H25}
  \foreach \j in {O3,O2,O4}
    \draw[line] (\i) -- (\j);

\node[draw=black, dashed, thick, fit=(I1) (I3) (H11) (H15) (O1), inner sep=10pt, label=above:{Feedforward ANN}] {};

\end{tikzpicture}
\caption{DML twin-network architecture.}
\label{fig:dml}
\end{figure}

Figure~\ref{fig:dml} illustrates the architectures considered: a standard feedforward ANN and its corresponding DML twin-network extension. When the integration domain is bounded, the endpoints $(a,b)$ are included as additional inputs, enabling the surrogate to generalize across domains.  

\section{Methodology} \label{sec:methodology}

Building on the principles established in Section~\ref{sec:preliminaries}, this section presents the methodological framework adopted in this work, detailing  the construction of training datasets based on single–simulation MC labels, and an illustrative example that anticipates the methodology employed in the numerical experiments.

\subsection{Generating Training Labels} \label{sec:gen_train_labels}

In order to train an ANN surrogate capable of approximating parametric integrals efficiently, we first need to construct an appropriate training dataset. Here, we adopt a uniform sampling scheme that generates unbiased single-realization MC targets, which serve as noisy yet statistically valid training labels. 

The procedure is as follows:
\begin{itemize}
\item Sample the inputs
\[
a^{(j)} \sim \mathcal{U}(a_{\min},a_{\max}), \qquad
b^{(j)} \sim \mathcal{U}(b_{\min},b_{\max}), \qquad
\boldsymbol{\theta}^{(j)} \sim \mathcal{U}(\boldsymbol{\theta}_{\min},\boldsymbol{\theta}_{\max}),
\]
subject to the constraint $a^{(j)} < b^{(j)}$.
\item Draw an auxiliary point uniformly in $(a^{(j)},b^{(j)})$,
\[
x^{(j)} \sim \mathcal{U}(a^{(j)},b^{(j)}),
\qquad \text{equivalently,} \qquad
x^{(j)} = a^{(j)} + (b^{(j)} - a^{(j)})\,u^{(j)}, \quad u^{(j)} \sim \mathcal{U}(0,1).
\]
\item Compute the output label using the unbiased single–simulation MC estimator,
\[
\hat{y}^{(j)} = (b^{(j)} - a^{(j)})\, f(x^{(j)};\boldsymbol{\theta}^{(j)}).
\]
\item Compute its gradient with respect to the full input vector $\boldsymbol{\hat{\theta}} = (a, b, \boldsymbol{\theta})$,
\[
\nabla_{\!\boldsymbol{\hat{\theta}}}\hat{y}^{(j)} 
= \nabla_{\!\boldsymbol{\hat{\theta}}}\!\big[(b^{(j)} - a^{(j)})\, f(x^{(j)};\boldsymbol{\theta}^{(j)})\big]
= \nabla_{\!\boldsymbol{\hat{\theta}}}(b^{(j)} - a^{(j)})\, f(x^{(j)};\boldsymbol{\theta}^{(j)}) 
+ (b^{(j)} - a^{(j)})\, \nabla_{\!\boldsymbol{\hat{\theta}}} f(x^{(j)};\boldsymbol{\theta}^{(j)}).
\]
\end{itemize}

Thus, we can construct a training dataset consisting of independently and randomly sampled triples $(\boldsymbol{\hat{\theta}}^{(j)}, \hat{y}^{(j)}, \nabla_{\!\boldsymbol{\hat{\theta}}} \hat{y}^{(j)})$, $j = 1,\ldots,J$. Furthermore, a separate testing dataset is generated for validation, where the MSE is evaluated against the ground-truth target value
\[
y^{\mathrm{true}} = I(\boldsymbol{\hat{\theta}}).
\]

\begin{remark} \label{rem:methodology}
The uniform sampling scheme above admits a natural generalization. Specifically, the auxiliary 
sample $x^{(j)}$ does not need to be restricted to $\mathcal{U}(a^{(j)},b^{(j)})$, but it may instead be drawn from a broader family of distributions. For instance,
\[
x^{(j)} \sim \nu,
\]
where $\nu$ is any probability distribution with support contained in $\mathbb{R}$. In this case, training labels are constructed as
\[
\hat{y}^{(j)} = f(x^{(j)};\boldsymbol{\theta}^{(j)}), 
\qquad 
\nabla_{\!\boldsymbol{\theta}}\hat{y}^{(j)} = \nabla_{\!\boldsymbol{\theta}} f(x^{(j)};\boldsymbol{\theta}^{(j)}).
\]
\end{remark}

\begin{remark}
In practice, DML generally improves the learning efficiency. However, there exits particular situations where it might be not always effective:
\begin{itemize}
\item If the integrand $f$ does not depend on all elements of $\boldsymbol{\theta}$, then no informative derivatives with respect to $\boldsymbol{\theta}$ appear, and the differential component of the training process degenerates.

\item If $\nabla_{\!\boldsymbol{\theta}}\hat{y}$ is non-smooth (for instance, due to asymptotes or singular behaviour), then it may fail to exist or may become numerically unstable in the neighbourhood of such points.
\end{itemize}
\end{remark}

\subsection{Illustrative Example}

To motivate the proposed approach, we consider the single integral
\[
\mathrm{I} = \int_0^{\pi} \cos(x)\,dx = \sin(\pi).
\]

\begin{figure}[t!]
\begin{center}
\begin{tikzpicture}[scale=2.0]
  
  \draw[->] (0,-1.1) -- (0,1.3) node[above]{$y$};
  \draw[->] (0,0) -- (3.5,0) node[right]{$x$};

  \draw[thick,blue,line width=2.0pt,domain=0:3.1416,samples=300] 
    plot(\x,{cos(deg(\x))}) node[right]{$y=\cos(x)$};

  \fill[blue,opacity=0.1] 
    plot[domain=0:3.1416,samples=300] (\x,{cos(deg(\x))}) -- (3.1416,0) -- (0,0) -- cycle;

  \foreach \i in {1,...,50} {
      \pgfmathsetmacro{\x}{3.1416*rnd}  
      \pgfmathsetmacro{\y}{cos(deg(\x))}
      \fill[red,opacity=0.8] (\x,\y) circle(0.8pt);
  }
\end{tikzpicture}
\end{center}
\caption{Uniform MC sampling for the integral of $\cos(x)$ over $[0,\pi]$.}
\label{fig:mc_cos}
\end{figure}
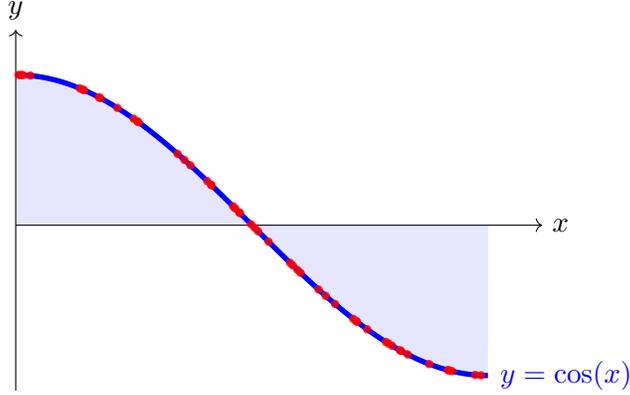

A plain MC method requires \(N\) independent simulations to approximate this integral once, as shown in Figure~\ref{fig:mc_cos}, given a prescribed accuracy of order $N^{-1/2}$. The scattered red points indicate evaluations of the integrand over uniformly generated realizations.

We now consider the parametric version, i.e.,
\[
\mathrm{I}(b) = \int_0^b \cos(x)\,\mathrm{d}x = \sin(b), \qquad b \in (0,\pi].
\]

With plain MC, estimating the integral $\mathrm{I}(b)$ for $M$ distinct $b$ values requires a total of $M \times N$ simulations, since each $b$ must be treated independently. In contrast, an ANN surrogate requires generating a single training dataset of size $J$ once. After training, the model $\widehat{\mathrm{I}}(b;\mathbf{w})$ can approximate $\mathrm{I}(b)$ for any $b$ within the considered interval at negligible marginal cost. Thus, whereas MC incurs a fresh sampling cost of $N$ per parameter value, the surrogate leverages a fixed dataset of size $J$ to approximate the entire parametric map. This ability to reuse training data, rather than resampling for each new parameter, is the central efficiency gain motivating our approach.

We next compare two models:
\begin{enumerate}
\item A baseline feedforward ANN, trained solely on function values, i.e., with $\vartheta = 0$ in \eqref{eq:combined_loss}.
\item A DML twin-network, trained via the loss~\eqref{eq:combined_loss} with $\vartheta \in (0, 1)$.
\end{enumerate}

The training data $(b^{(j)}, \hat{y}^{(j)}, \nabla_{\!b }\hat{y}^{(j)})$, $j = 1,\ldots,J$ are generated as
\[
\hat{y}^{(j)} = b^{(j)} \cos(x^{(j)}), 
\qquad 
x^{(j)} = b^{(j)} z^{(j)}, \quad z^{(j)} \sim \mathcal{U}(0,1),
\]
together with their parameter derivatives,
\[
\frac{\partial \hat{y}^{(j)}}{\partial b} 
= \cos(x^{(j)}) - b^{(j)} z^{(j)} \sin(x^{(j)}).
\]

\begin{figure}[t!]
\centering
\begin{subfigure}{0.48\textwidth}
  \centering
  \includegraphics[width=\linewidth]{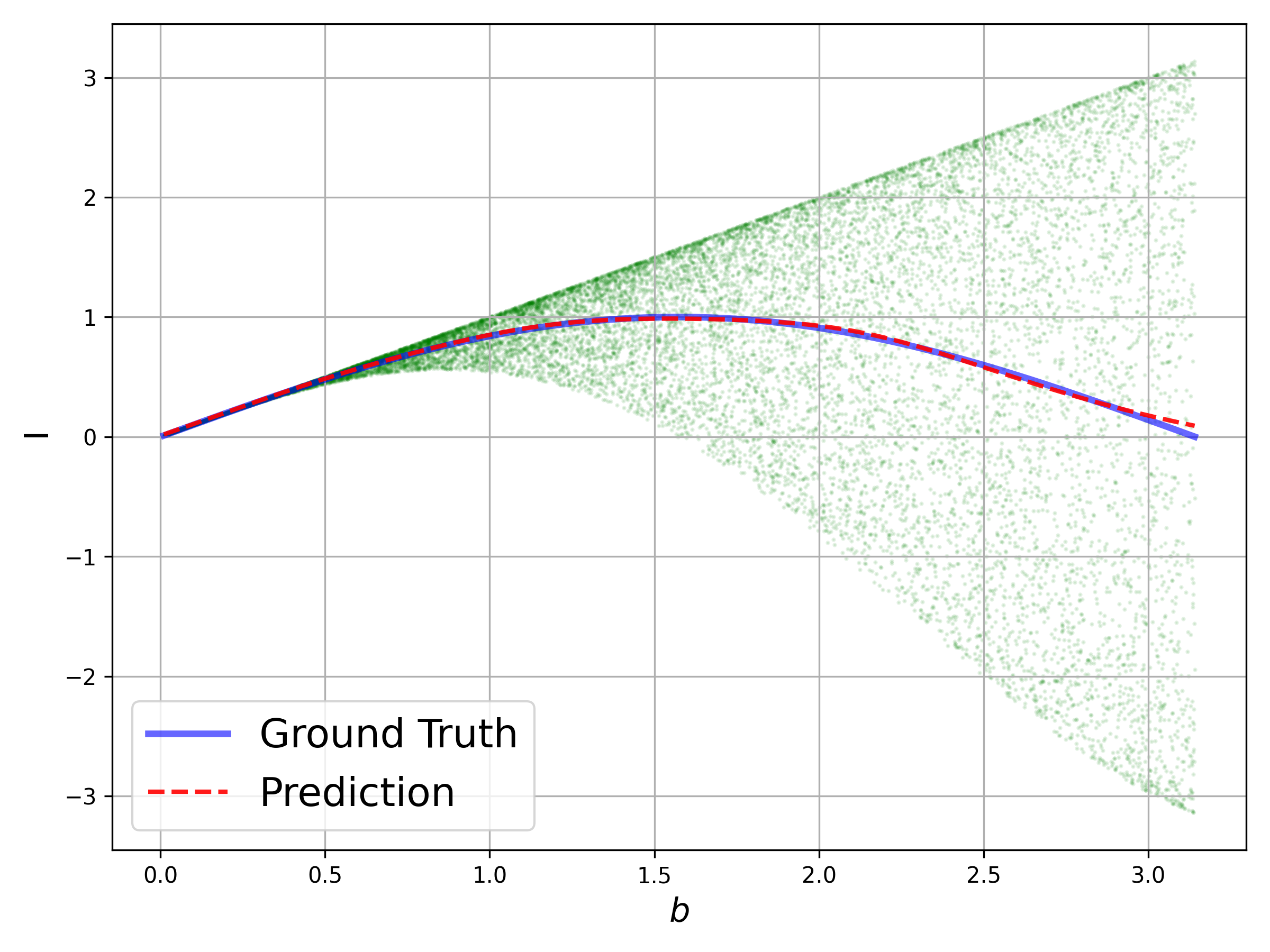}
  \label{fig:ann_cos_pred}
\end{subfigure}
\hfill
\begin{subfigure}{0.48\textwidth}
  \centering
  \includegraphics[width=\linewidth]{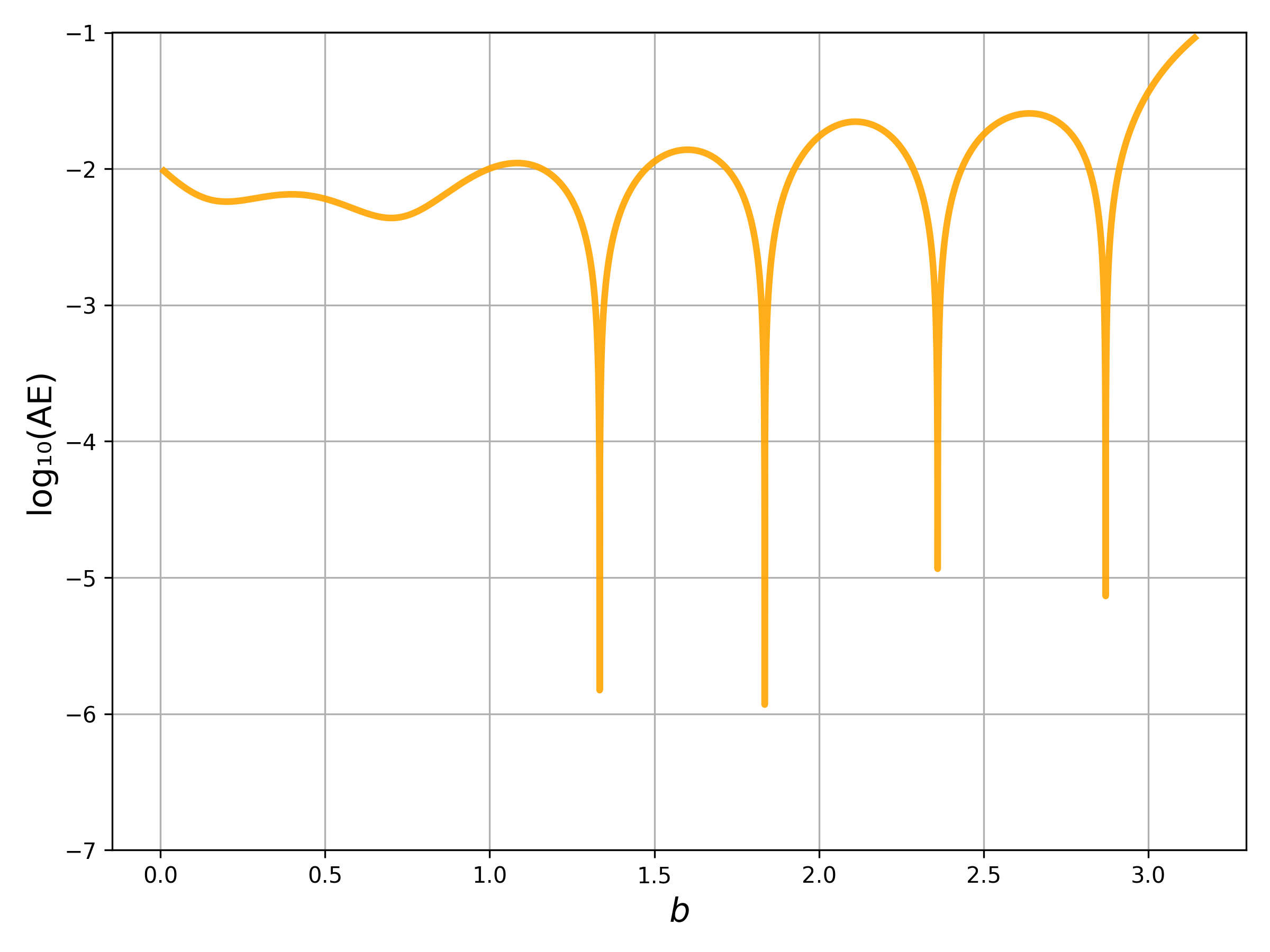}
  \label{fig:ann_cos_error}
\end{subfigure}

\vspace{0.2em}

\begin{subfigure}{0.48\textwidth}
  \centering
  \includegraphics[width=\linewidth]{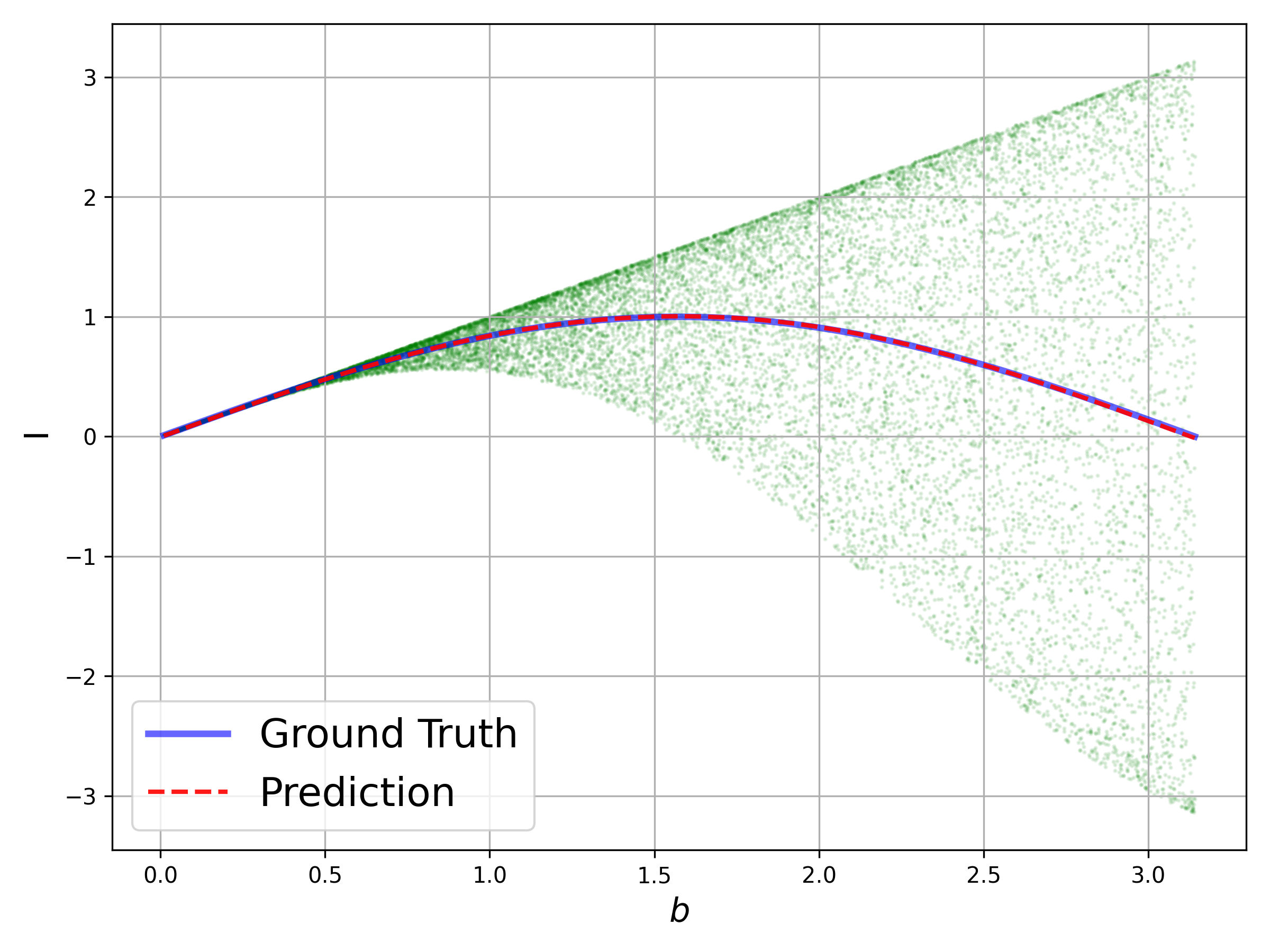}
  \label{fig:dml_cos_pred}
\end{subfigure}
\hfill
\begin{subfigure}{0.48\textwidth}
  \centering
  \includegraphics[width=\linewidth]{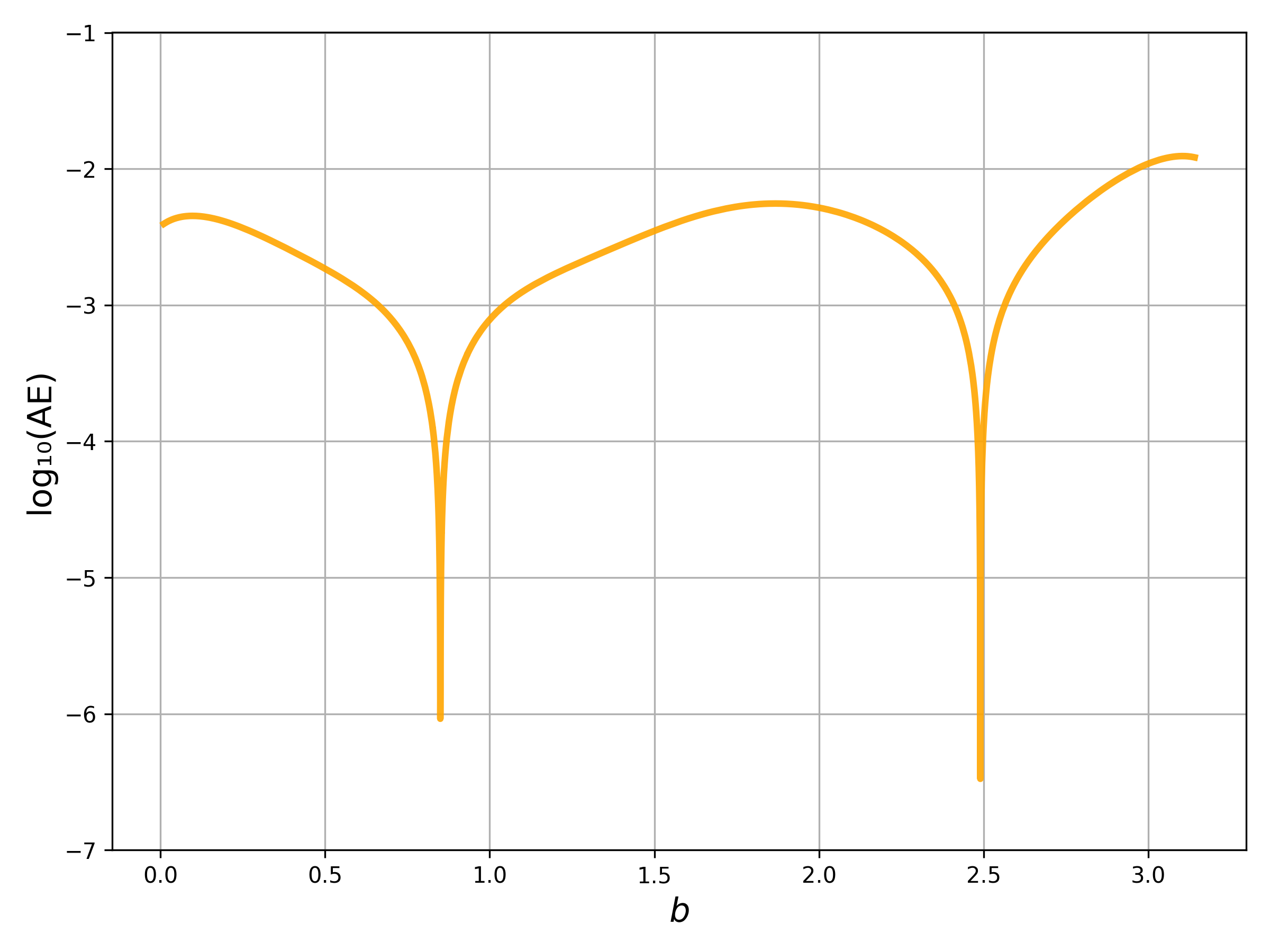}
  \label{fig:dml_cos_error}
\end{subfigure}

\caption{Predictions vs.\ analytical solutions (left column) and log-scale absolute errors (right column) for ANN (top row) and DML (bottom row) for the illustrative example with \( b \in [0.01, \pi] \).}
\label{fig:all_cos}
\end{figure}

Figure~\ref{fig:all_cos} compares ANN and DML predictions against their analytical solutions for $J = 2^{16}$, and reports the resulting absolute error (AE). The baseline ANN captures the general trend but exhibits noticeable errors, particularly near the upper end of the domain where the derivative is large. By contrast, the DML model closely follows the exact solution across the entire domain, with substantially reduced absolute error. Note that the green markers correspond to the noisy labels used during training.

\begin{figure}[t!]
\centering
\includegraphics[width=0.5\textwidth]{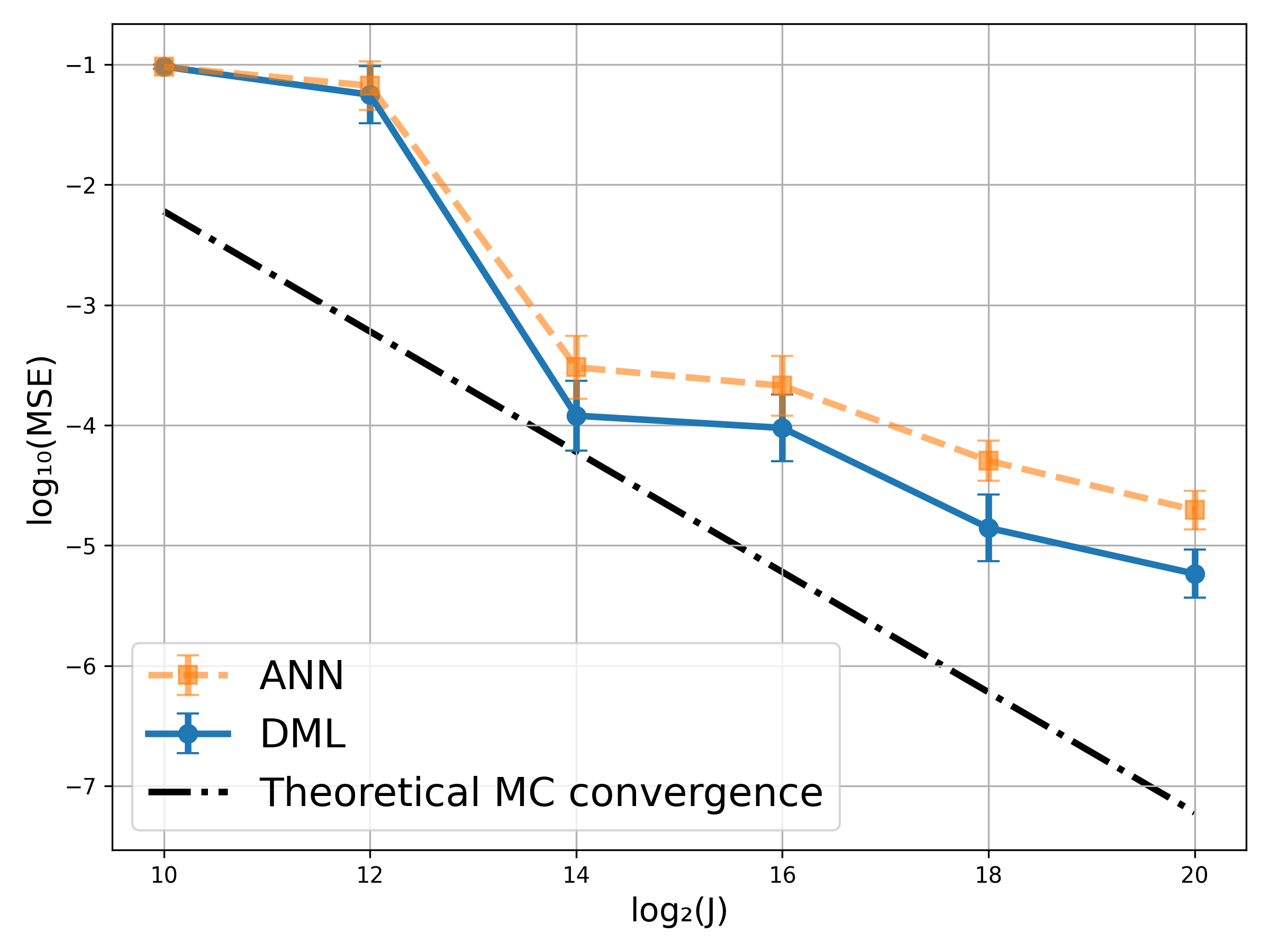}
\caption{MSE vs.\ training set size for the illustrative example with \( b \in [0.01,\pi] \).}
\label{fig:error_comparison_cos}
\end{figure}

Additionally, Figure~\ref{fig:error_comparison_cos} shows the MSE decay with respect to the training set size $J$. The DML model achieves faster convergence and maintains uniform accuracy across the domain, demonstrating the sample efficiency gains.

This toy example already anticipates some of the major benefits due to the use of DML in this context, such that the robust approximation in profound curvature regions or boundaries, as well as consistent lower errors as the dataset size increases.

The established experimental setup in the example (comparing baseline ANNs against the DML model) will serve as the reference procedure for the experiments presented in the following section.

\section{Numerical Results}
\label{sec:results}

This section reports a series of numerical experiments designed to assess the performance of the proposed DML framework against standard feedforward ANNs when approximating parametric integrals. The analysis focuses on accuracy, scalability, and sample efficiency across distinct classes of problems: statistical functionals (moments and CDFs), functional approximation via Chebyshev expansions, and integrals arising naturally from differential equations. Each subsection provides visual comparisons between DML and classical ANN surrogates, followed by a concluding discussion that incorporates the main findings. 

All the codes were implemented in \texttt{Python 3.9.5} using \texttt{TensorFlow 2.12.1} and run in a system equipped with Intel Core i7-4720HQ 2.6GHz CPU processors, RAM memory of 16GB and a GPU Nvidia Tesla V100. The hyperparameter optimization was performed through the \texttt{Keras-Tuner 1.4.8} library~\cite{keras_tuner}. Unless otherwise indicated, both models were trained using the data construction procedure described in Section \ref{sec:gen_train_labels}, applying a preprocessing based on data standardization as suggested in \cite{huge2020}. As optimizer, we choose the commonly utilized Adam, considering a learning rate decaying quadratically from $10^{-2}$ to $10^{-5}$. The default architecture consists of four hidden layers of 64 neurons. Training proceeded for 128 epochs with a mini-batch size of 1024 samples. Moreover, for the convergence tests, the MSE corresponding to each training set size $J$ is calculated as the mean over 10 trials.

\subsection{Parametric Statistics} \label{sec:param_statistics}

A primary application of the proposed methodology is the approximation of parametric statistics (such as moments and CDFs) which are central in probability theory, statistical inference, and stochastic modelling.

\subsubsection{Parametric Moments}

We begin by approximating parametric moments. These quantities encapsulate essential features of probability distributions such as mean, variance, skewness, and kurtosis. For a random variable $X$ with measure $\nu$, the $m$-th moment is defined as
\begin{equation*}
    \mathbb{E}_{\nu}[X^{m}] = \int_{-\infty}^{+\infty} x^{m}\,\nu(dx), \qquad m \in \mathbb{R}.
\end{equation*}

\subsubsection*{Experiment 1: Lognormal Distribution}

We first examine the case where $X \sim \mathrm{LogNormal}(\mu, \sigma^{2})$, with parameter
vector $\boldsymbol{\theta} = (m, \mu, \sigma)$. For this distribution, the fractional moment admits a closed-form analytical expression,
\begin{equation}
    \mathrm{M}(\boldsymbol{\theta}) = \mathbb{E}[X^{m}]
    = \exp\!\left(m\mu + \tfrac{1}{2} m^{2}\sigma^{2}\right),
    \label{eq:lognormal_moment}
\end{equation}
which serves as an exact reference for assessing the surrogate's accuracy.

To construct the training data, we draw independent standard normal samples $z^{(j)} \sim \mathcal{N}(0,1)$ and define the corresponding single-realization MC labels according to Remark~\ref{rem:methodology}, i.e.,
\begin{equation*}
    \hat{y}^{(j)} = \exp\!\big(m^{(j)}(\mu^{(j)} + \sigma^{(j)} z^{(j)})\big),
    \qquad j = 1,\ldots,J,
\end{equation*}
which represent unbiased estimators of \eqref{eq:lognormal_moment} under the sampling scheme. The associated parameter differentials, used in the loss \eqref{eq:combined_loss}, are given by
\begin{align*}
    \frac{\partial \hat{y}^{(j)}}{\partial m}
        &= \big(\mu^{(j)} + \sigma^{(j)} z^{(j)}\big)\, \hat{y}^{(j)}, &
    \frac{\partial \hat{y}^{(j)}}{\partial \mu}
        &= m^{(j)} \hat{y}^{(j)}, &
    \frac{\partial \hat{y}^{(j)}}{\partial \sigma}
        &= m^{(j)} z^{(j)} \hat{y}^{(j)}.
\end{align*}

\begin{figure}[t!]
\centering
\begin{subfigure}[t]{0.48\textwidth}
    \centering
    \includegraphics[width=\linewidth]{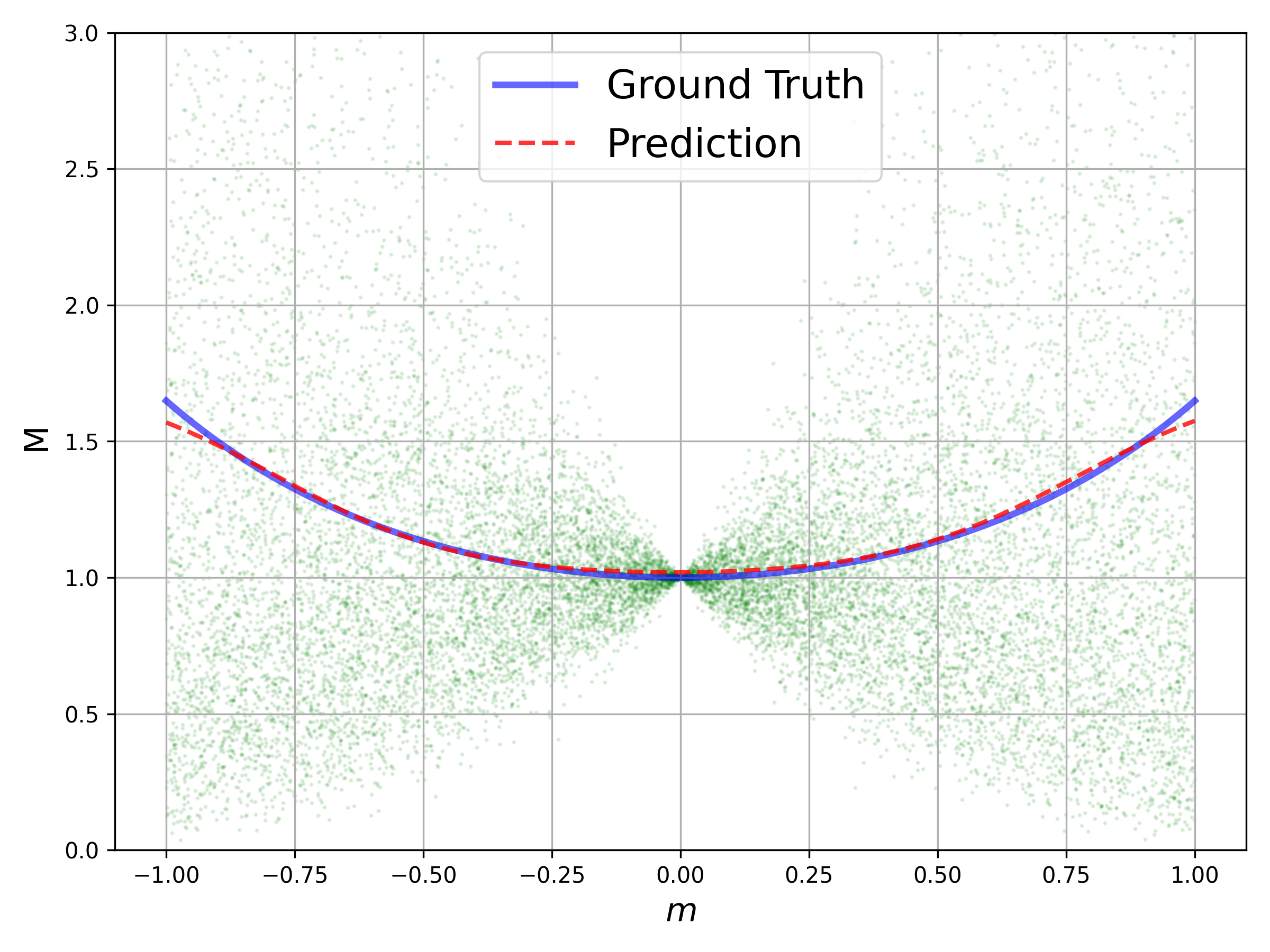}
\end{subfigure}
\hfill
\begin{subfigure}[t]{0.48\textwidth}
    \centering
    \includegraphics[width=\linewidth]{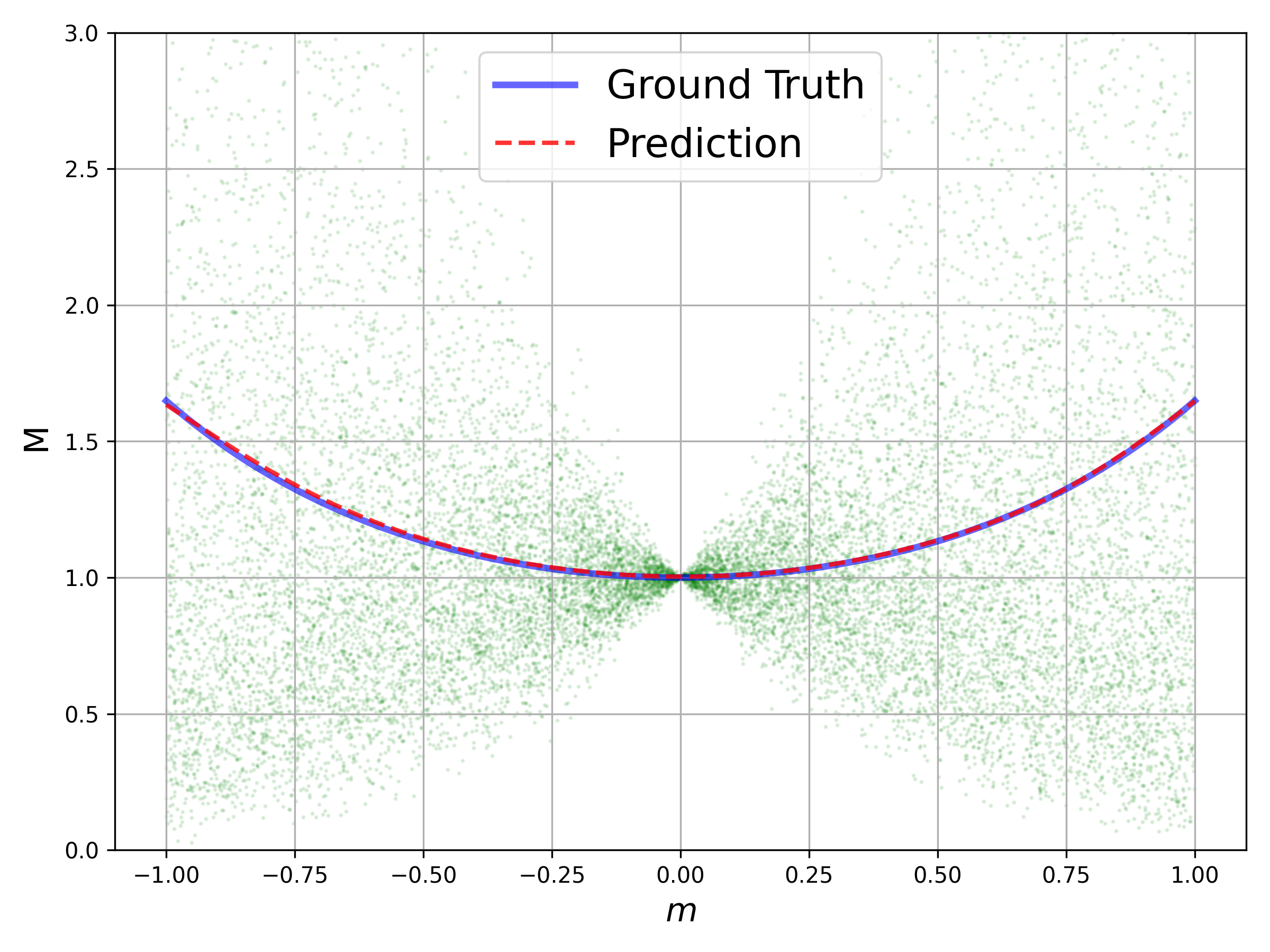}
\end{subfigure}
\caption{Predictions vs.\ analytical moments for ANN (left) and DML (right) for $X \sim \mathrm{LogNormal}(0,1)$ and $m \in [-1,1]$.}
\label{fig:lognormal_moments}
\end{figure}

\begin{figure}[t!]
\centering
\begin{subfigure}{0.48\textwidth}
    \centering
    \includegraphics[width=\linewidth]{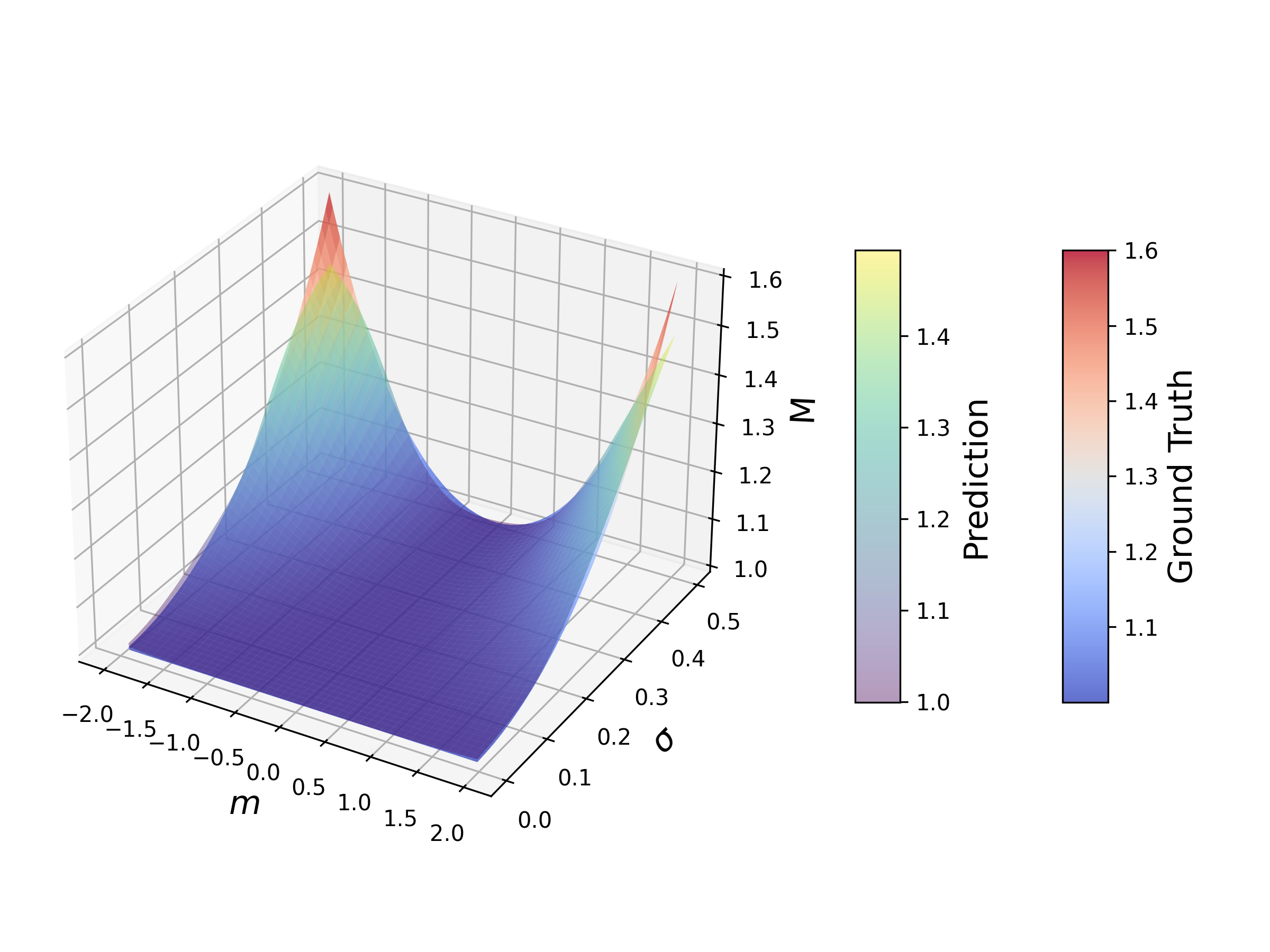}
    \label{fig:ann_lognormal_pred_2i}
\end{subfigure}
\hfill
\begin{subfigure}{0.48\textwidth}
    \centering
    \includegraphics[width=\linewidth]{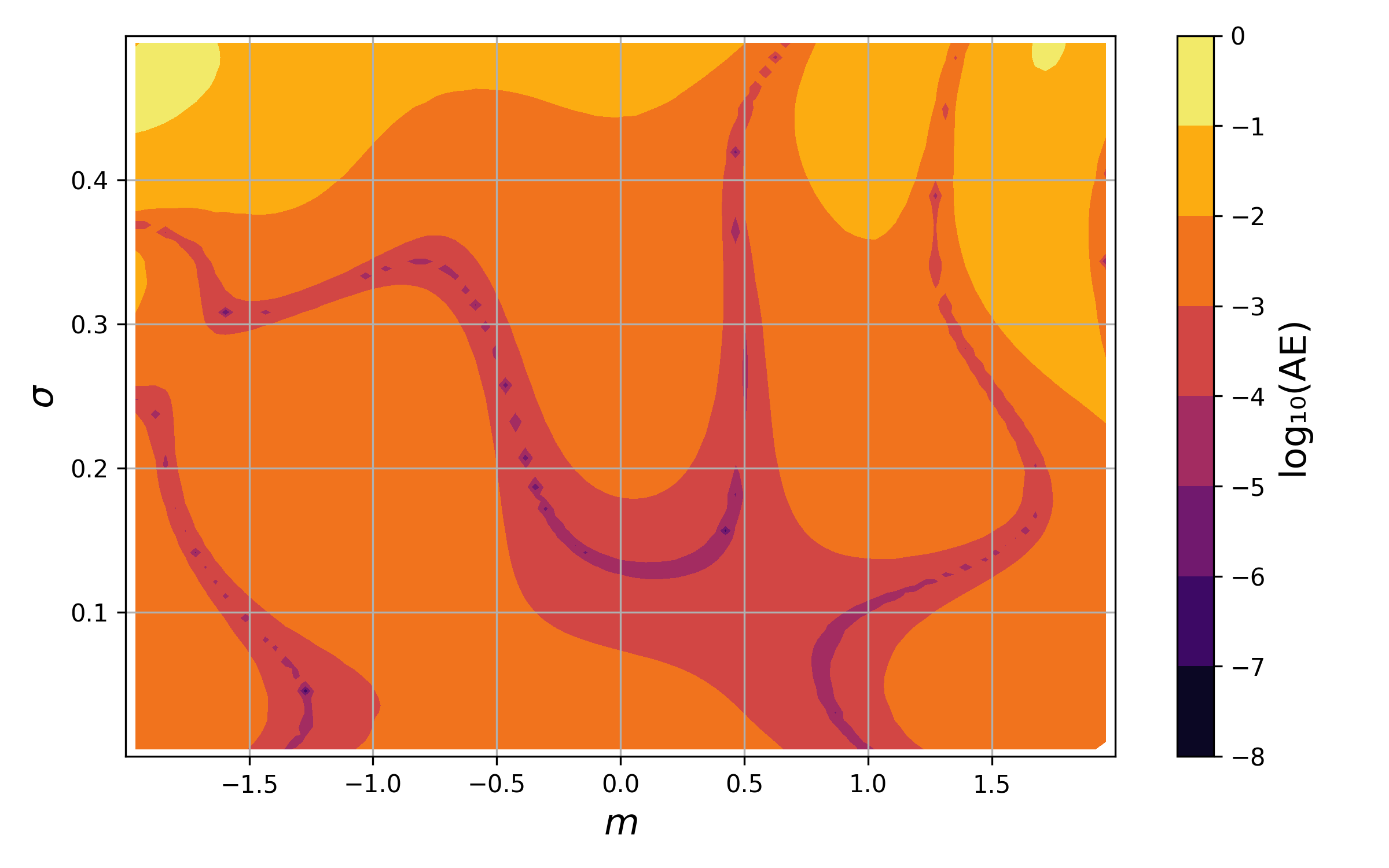}
    \label{fig:ann_lognormal_error_2i}
\end{subfigure}
\vspace{0.2em}
\begin{subfigure}{0.48\textwidth}
    \centering
    \includegraphics[width=\linewidth]{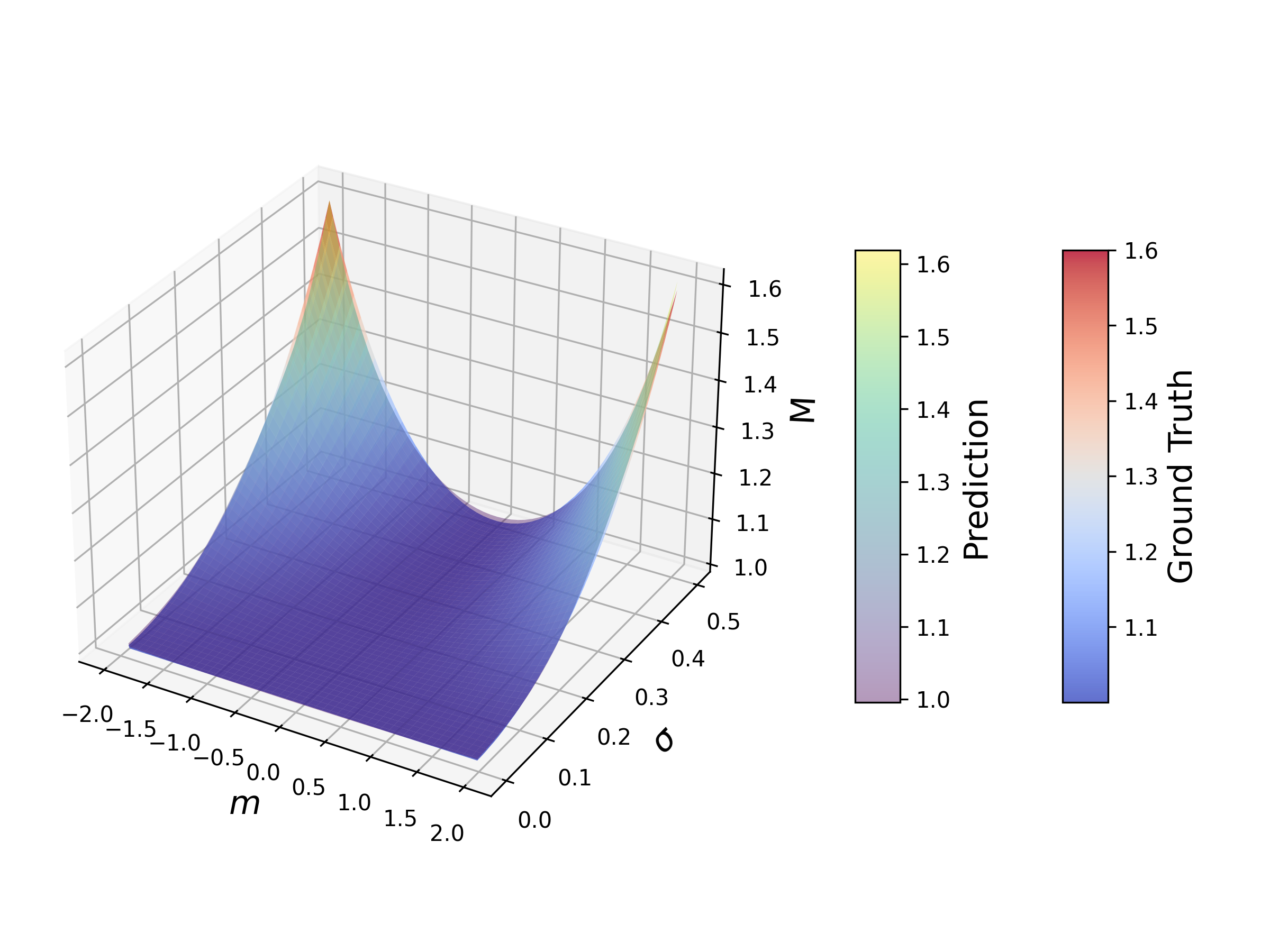}
    \label{fig:dml_lognormal_pred_2i}
\end{subfigure}
\hfill
\begin{subfigure}{0.48\textwidth}
    \centering
    \includegraphics[width=\linewidth]{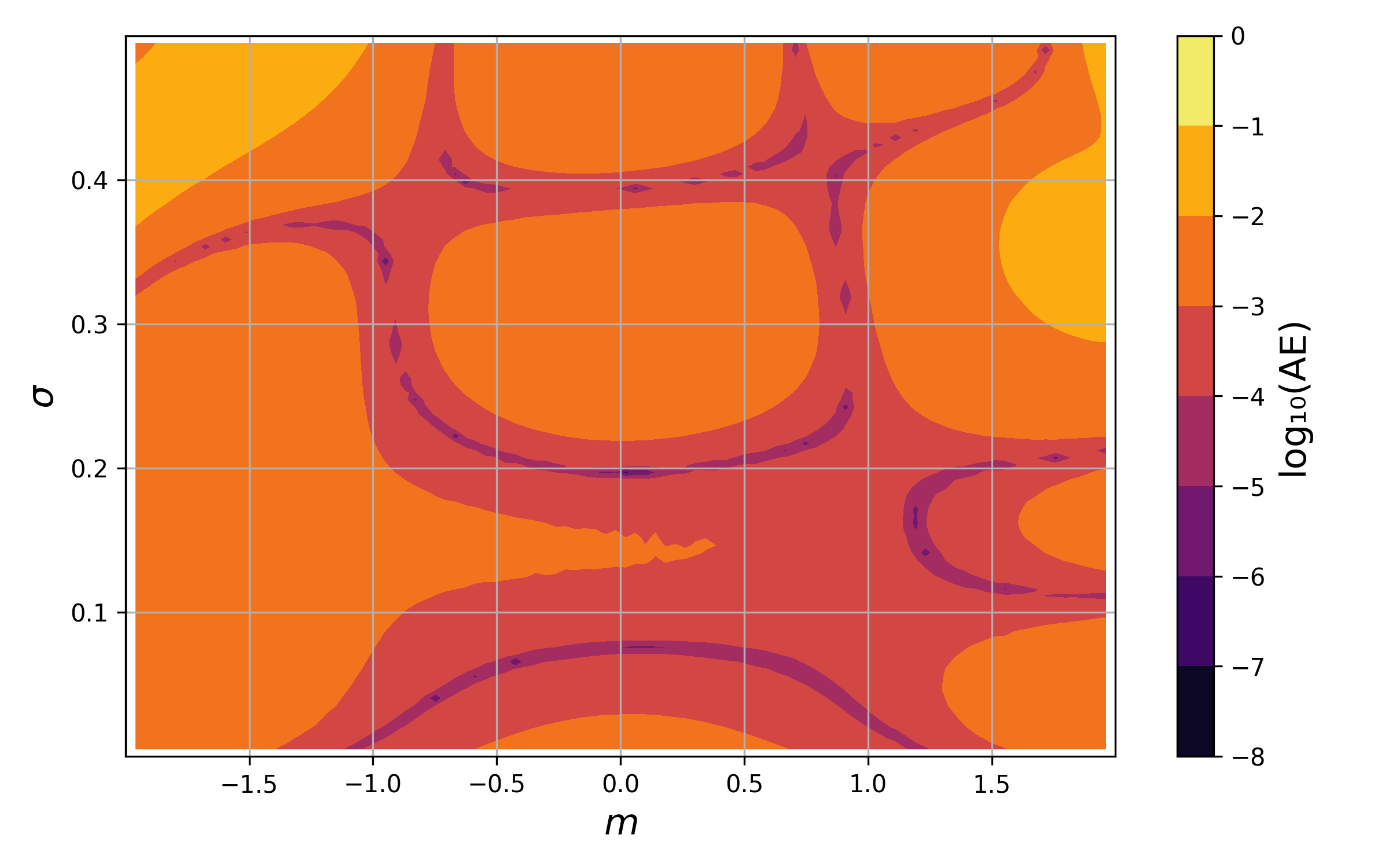}
    \label{fig:dml_lognormal_error_2i}
\end{subfigure}
\caption{Predictions vs. analytical moments (left) and log-scale absolute errors (right) for ANN (top row) and DML (bottom row) for $X \sim \mathrm{LogNormal}(0,\sigma^2)$, $m \in [-2,2]$ and $\sigma \in [0,0.5]$.}
\label{fig:lognormal_moments_2D}
\end{figure}

\begin{figure}[t!]
\centering
\begin{subfigure}[t]{0.48\textwidth}
    \centering
    \includegraphics[width=\linewidth]{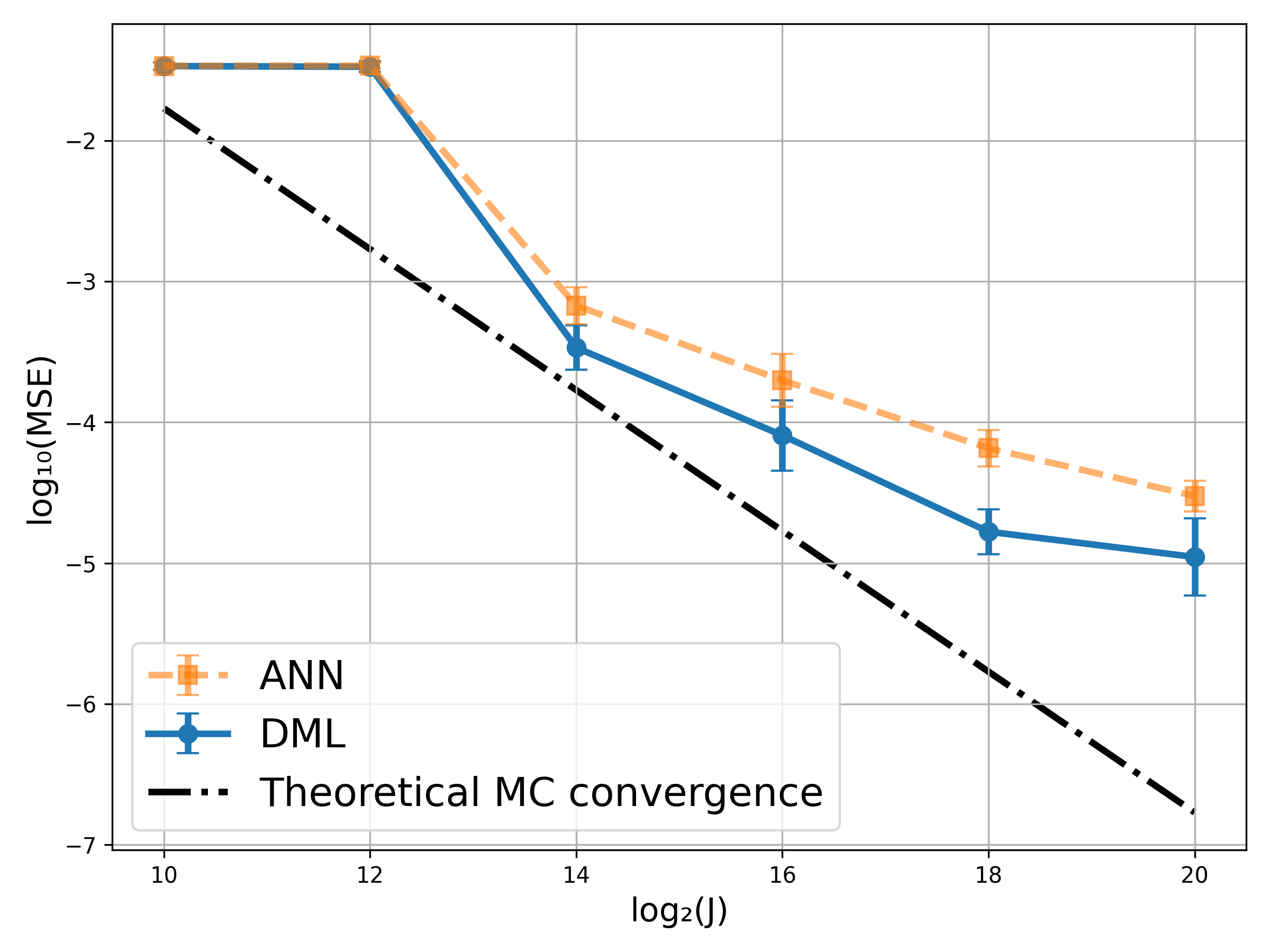}
    \caption*{$X \sim \mathrm{LogNormal}(0,1)$, $m \in [-1,1]$}
\end{subfigure}
\hfill
\begin{subfigure}[t]{0.48\textwidth}
    \centering
    \includegraphics[width=\linewidth]{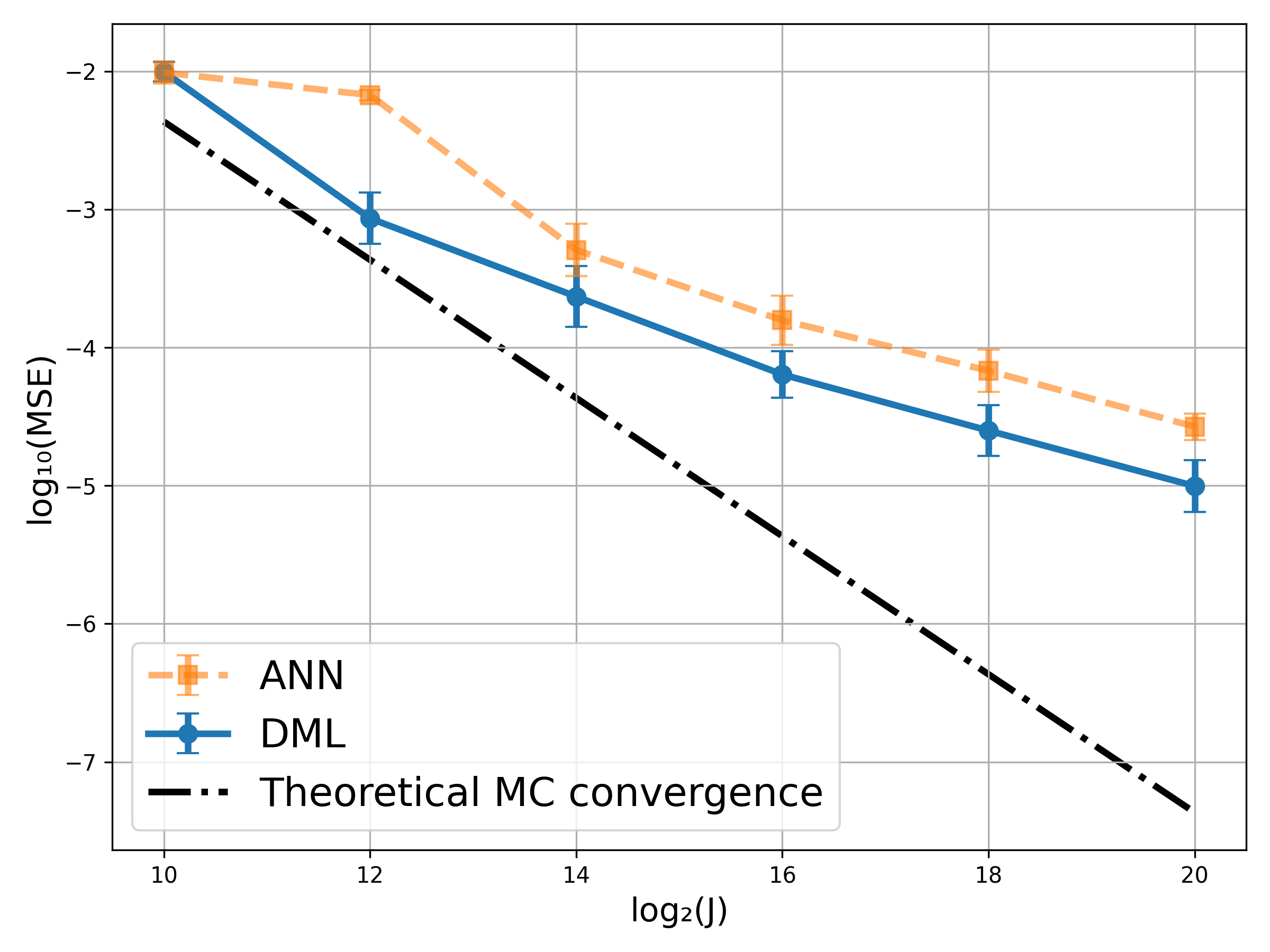}
    \caption*{$X \sim \mathrm{LogNormal}(0,\sigma^2)$, $m \in [-2,2]$, $\sigma \in [0,0.5]$}
\end{subfigure}
\caption{MSE vs.\ training set size for parametric moments.}
\label{fig:moments}
\end{figure}

In Figures~\ref{fig:lognormal_moments}–\ref{fig:moments}, we present the results obtained for the benchmark examples of parametric moments under the lognormal distribution for $J = 2^{16}$. Figure~\ref{fig:lognormal_moments} compares the predictions of the baseline ANN and the proposed DML framework for the one-input case $X \sim \mathrm{LogNormal}(0,1)$ with $m \in [-1,1]$, while Figure~\ref{fig:lognormal_moments_2D} extends the analysis to the two-input setting $X \sim \mathrm{LogNormal}(0,\sigma^2)$, $m \in [-2,2]$, and $\sigma \in [0,0.5]$. Finally, Figure~\ref{fig:moments} displays the MSE as a function of the training set size.

\subsubsection{Parametric Cumulative Distribution Functions}

We next consider the approximation of parametric CDFs. For a random variable $X$ with probability density function (PDF) $f_X(x;\boldsymbol{\theta})$, the CDF is defined by
\begin{equation*}
    \mathrm{F}(b;\boldsymbol{\theta}) = \int_{-\infty}^{b} f_X(x;\boldsymbol{\theta})\,\mathrm{d}x.
\end{equation*}

To illustrate the performance of the proposed ANN and DML frameworks, we consider two representative distribution families: the Chi-squared and the Normal Inverse Gaussian (NIG). Throughout this section, we denote by $\hat{\boldsymbol{\theta}} = (b,\boldsymbol{\theta})$ the input vector comprising the upper integration limit and distribution parameters.

\subsubsection*{Experiment 2: Chi-Squared Distribution}

We begin by considering $X \sim \chi^{2}(\theta)$, where $\theta > 0$ denotes the degrees of freedom. For this case, a closed-form expression of the CDF is available
\begin{equation}
    \mathrm{F}_{\chi^{2}}(b;\theta)
    = \int_{0}^{b} f_{\chi^{2}}(x;\theta)\,\mathrm{d}x
    = \frac{\gamma(\theta/2,\, b/2)}{\Gamma(\theta/2)},
    \label{eq:chi2_cdf}
\end{equation}
where $\gamma(\cdot,\cdot)$ is the lower incomplete gamma function, and the corresponding PDF is given by
\begin{equation*}
    f_{\chi^{2}}(x;\theta)
    = \frac{1}{2^{\theta/2}\Gamma(\theta/2)}
      \,x^{\frac{\theta}{2}-1} e^{-x/2},
    \qquad x>0.
\end{equation*}

The training samples $(\hat{\boldsymbol{\theta}}^{(j)}, \hat{y}^{(j)}, \nabla_{\!\hat{\boldsymbol{\theta}}}\hat{y}^{(j)})$, $j = 1,\ldots,J$, are constructed as single-realization MC labels
\begin{equation*}
    \hat{y}^{(j)} = b^{(j)} f_{\chi^{2}}\!\big(x^{(j)};\theta^{(j)}\big),
    \qquad
    x^{(j)} = b^{(j)} u^{(j)}, \quad u^{(j)} \sim \mathcal{U}(0,1),
\end{equation*}
which provides unbiased estimators of \eqref{eq:chi2_cdf}. The corresponding parameter differentials, required for the loss \eqref{eq:combined_loss}, are computed as
\begin{align*}
    \frac{\partial \hat{y}^{(j)}}{\partial b}
        &= \frac{\theta^{(j)} - x^{(j)}}{2}\,
           f_{\chi^{2}}\!\big(x^{(j)};\theta^{(j)}\big), &
    \frac{\partial \hat{y}^{(j)}}{\partial \theta}
        &= \hat{y}^{(j)}\!
           \left[
               \tfrac{1}{2}\ln x^{(j)}
               - \tfrac{1}{2}\ln 2
               - \tfrac{1}{2}\psi\!\left(\tfrac{\theta^{(j)}}{2}\right)
           \right],
\end{align*}
where $\psi$ denotes the digamma function.

\begin{figure}[t!]
\centering
\begin{subfigure}[t]{0.48\textwidth}
    \centering
    \includegraphics[width=\linewidth]{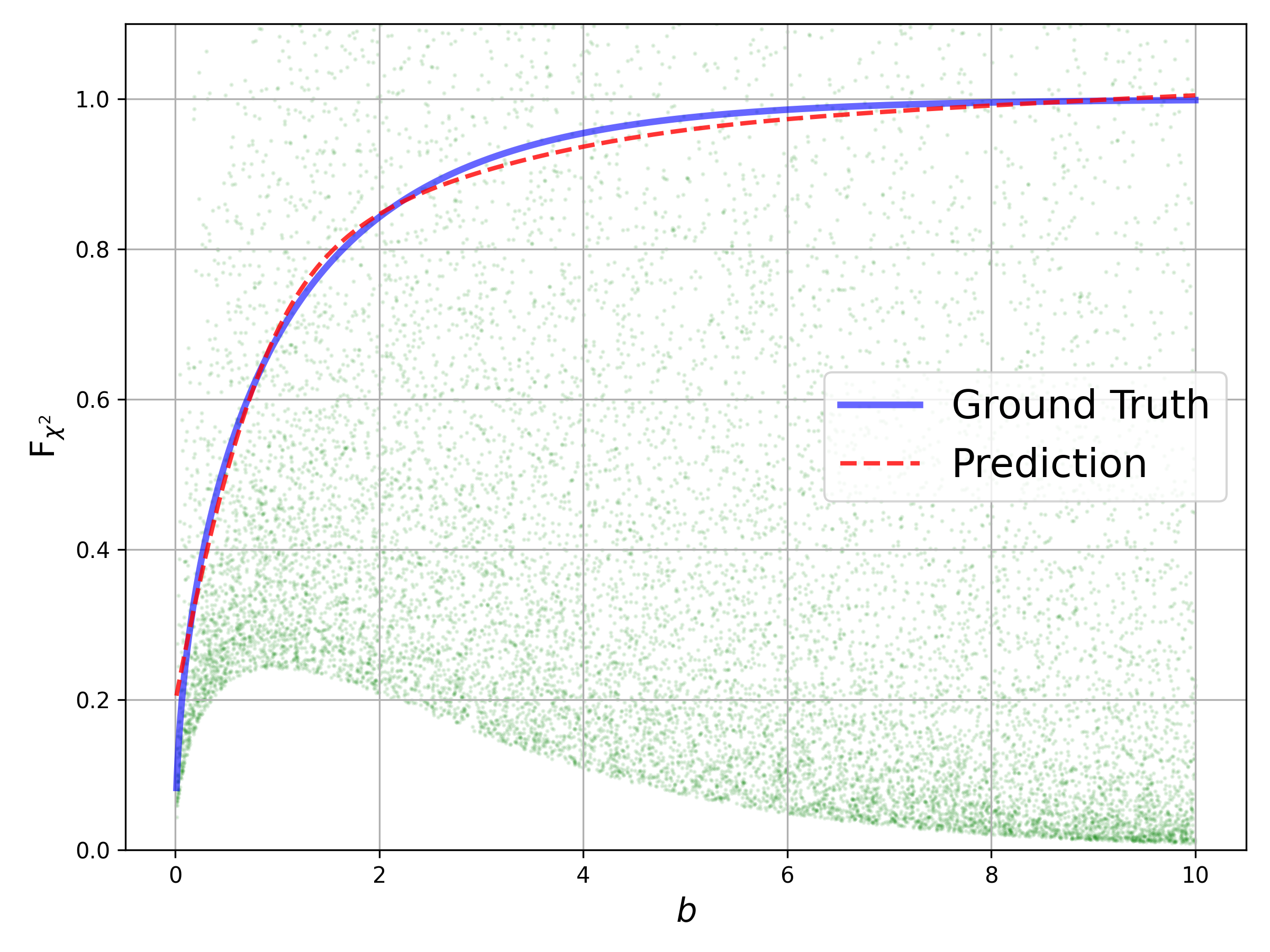}
\end{subfigure}
\hfill
\begin{subfigure}[t]{0.48\textwidth}
    \centering
    \includegraphics[width=\linewidth]{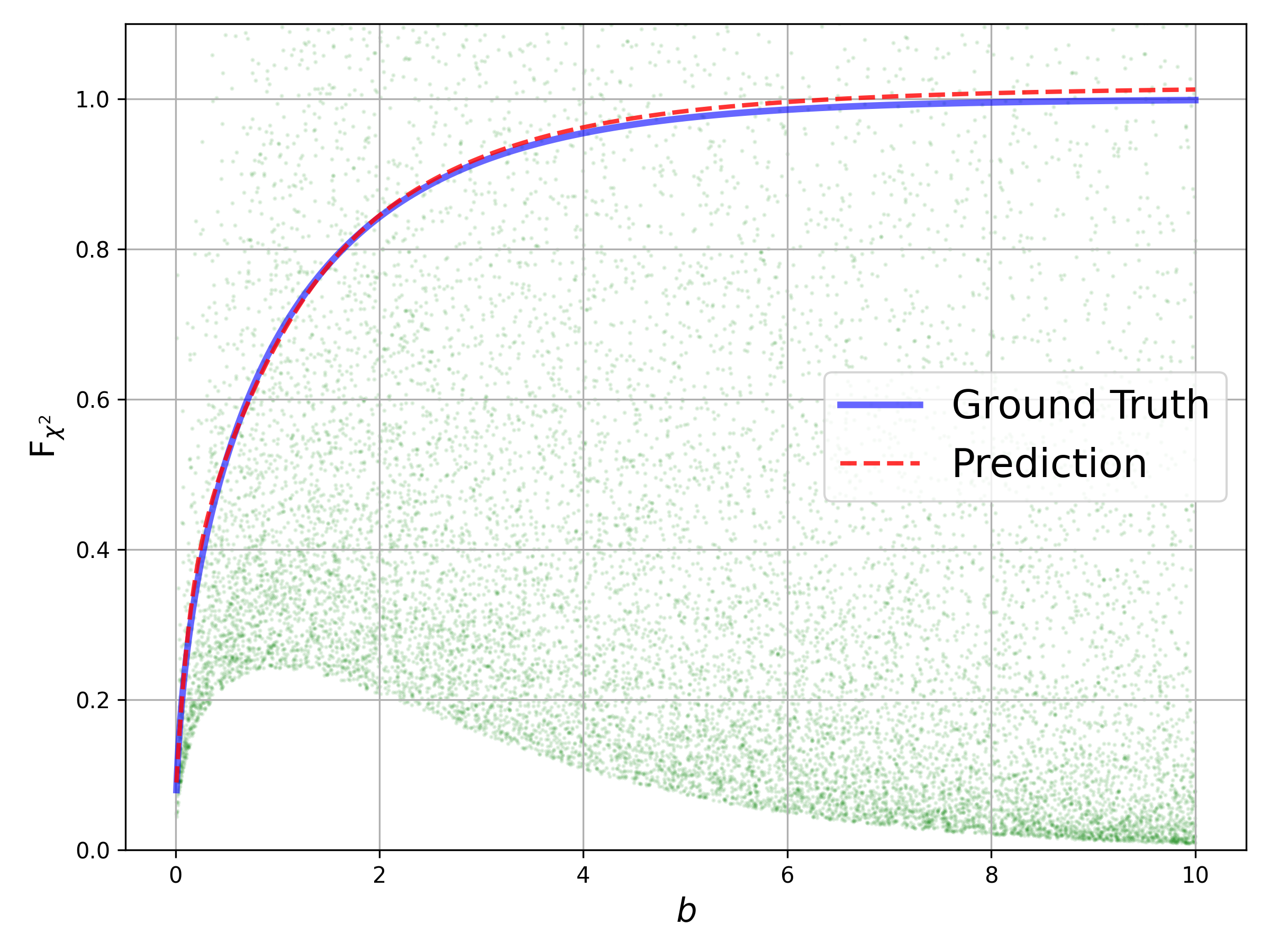}
\end{subfigure}
\caption{Predictions vs.\ analytical CDFs for ANN (left) and DML (right) for $X \sim \chi^2(1)$, $b \in [0.01,10]$.}
\label{fig:chi_CDFs}
\end{figure}

\begin{figure}[t!]
\centering
\begin{subfigure}{0.48\textwidth}
    \centering
    \includegraphics[width=\linewidth]{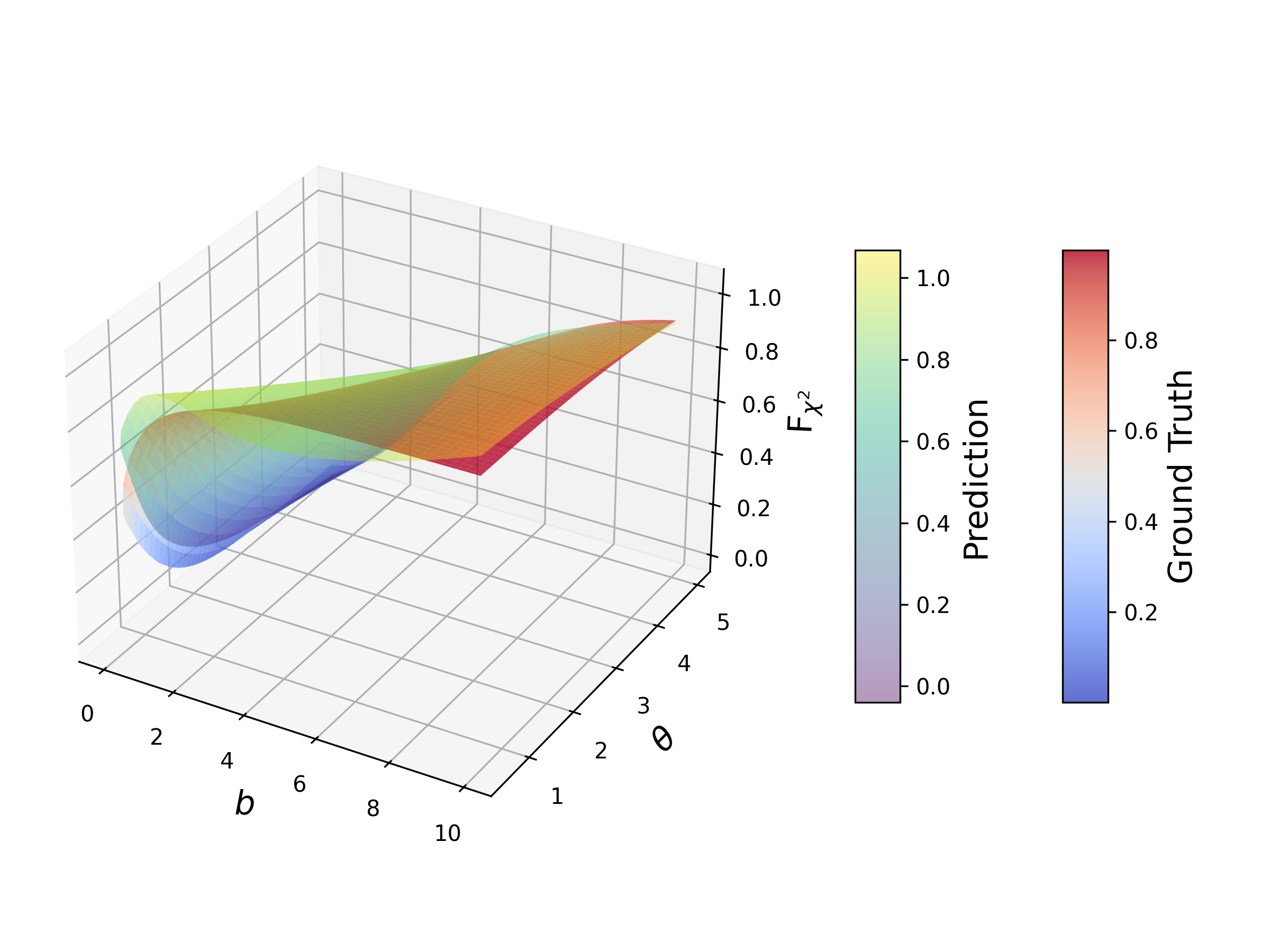}
    \label{fig:ann_chi_pred_2i}
\end{subfigure}
\hfill
\begin{subfigure}{0.48\textwidth}
    \centering
    \includegraphics[width=\linewidth]{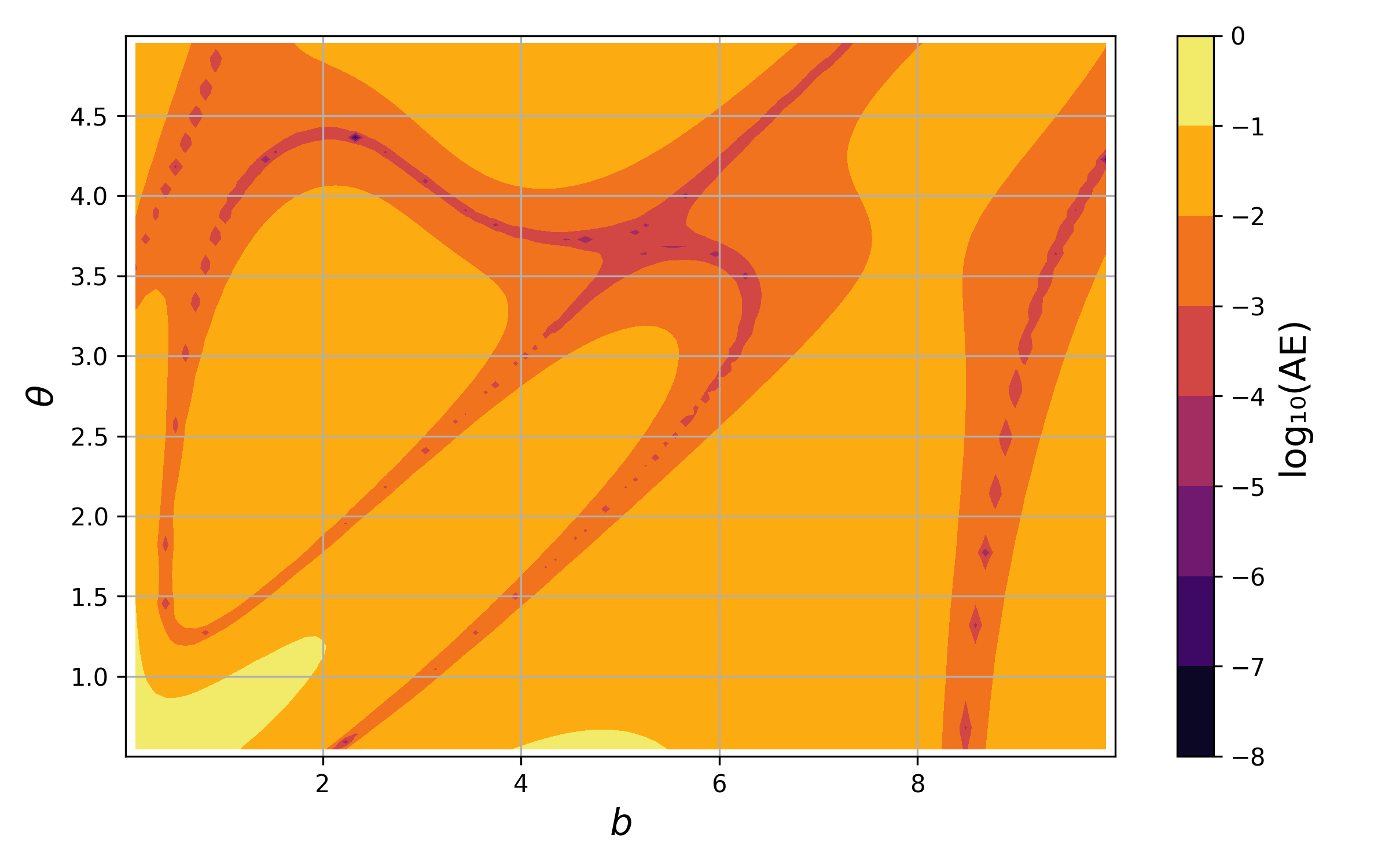}
    \label{fig:ann_chi_error_2i}
\end{subfigure}
\vspace{0.2em}
\begin{subfigure}{0.48\textwidth}
    \centering
    \includegraphics[width=\linewidth]{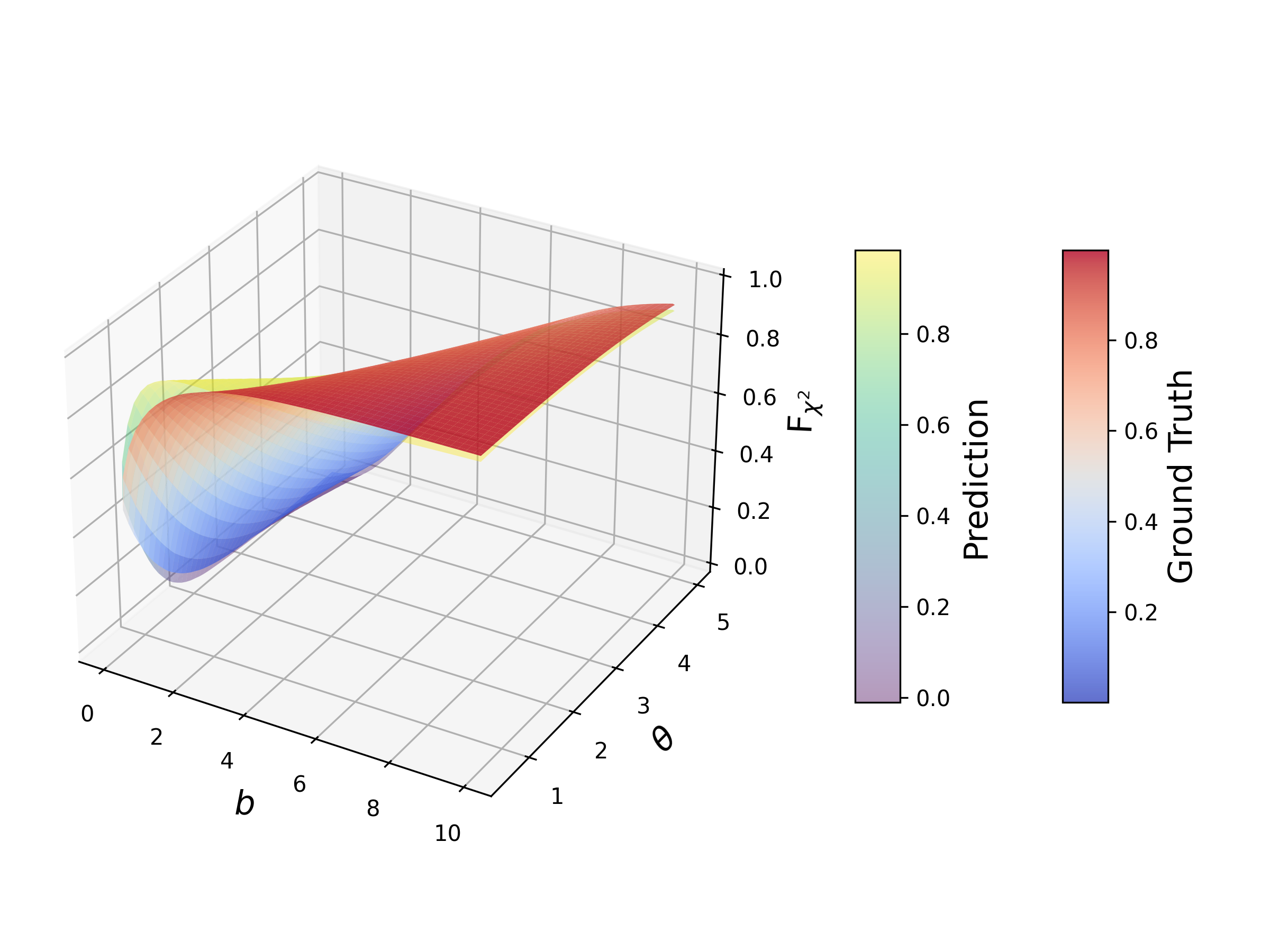}
    \label{fig:dml_chi_pred_2i}
\end{subfigure}
\begin{subfigure}{0.48\textwidth}
    \centering
    \includegraphics[width=\linewidth]{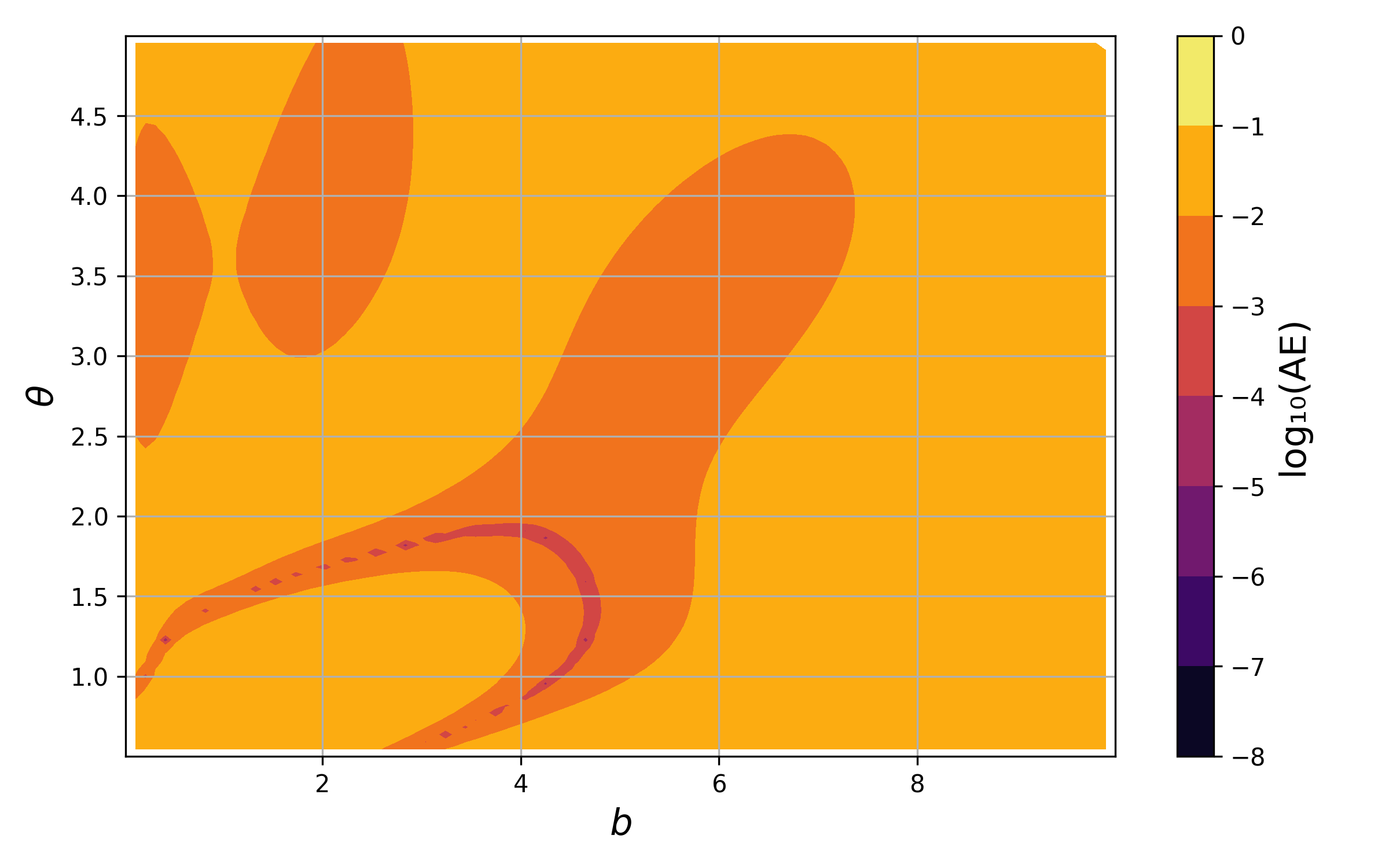}
    \label{fig:dml_chi_error_2i}
\end{subfigure}
\caption{Predictions vs.\ analytical CDFs (left) and log-scale absolute errors (right) for ANN (top row) and DML (bottom row) for $X \sim \chi^2(\theta)$, $b \in [0.01,10]$ and $\theta \in [0.5,5.0]$.}
\label{fig:chi_CDFs_2D}
\end{figure}

In Figures~\ref{fig:chi_CDFs}–\ref{fig:chi_CDFs_2D}, we present the results obtained for the benchmark examples of parametric CDFs under the chi-squared family for $J = 2^{16}$. Figure~\ref{fig:chi_CDFs} compares the predictions of the baseline ANN and the proposed DML model for the one-input case $X \sim \chi^2(1)$ with $b \in [0.01,10]$, while Figure~\ref{fig:chi_CDFs_2D} extends the analysis to the two-input setting $X \sim \chi^2(\theta)$, $b \in [0.01,10]$, and $\theta \in [0.5,5.0]$.

\subsubsection*{Experiment 3: Normal Inverse Gaussian Distribution}

We finally consider the NIG distribution, $X \sim \mathrm{NIG}(\boldsymbol{\theta})$, where $\boldsymbol{\theta} = (\alpha,\beta,\mu,\delta)$. The tail heaviness is governed by $\alpha>0$, while the skewness is controlled by
$\beta$ with $|\beta|<\alpha$. The parameters $\mu \in \mathbb{R}$ and $\delta > 0$ represent the location and scale, respectively. Let $\mathcal{X} = [a,b]$, the truncated CDF is defined as
\begin{equation}
    \mathrm{F}_{\mathrm{NIG}}(b;\boldsymbol{\theta})
    = \int_{a}^{b} f_{\mathrm{NIG}}(x;\boldsymbol{\theta})\,\mathrm{d}x,
    \label{eq:nig_cdf}
\end{equation}
where $f_{\mathrm{NIG}}$ denotes the PDF of the NIG distribution
\begin{equation}
    f_{\mathrm{NIG}}(x;\boldsymbol{\theta})
    = \frac{\alpha \delta}{\pi}
      \exp\!\big(\delta \tau + \beta (x-\mu)\big)
      \frac{K_{1}(\alpha \upsilon)}{\upsilon},
    \qquad
    \tau = \sqrt{\alpha^{2} - \beta^{2}},\quad
    \upsilon = \sqrt{\delta^{2} + (x-\mu)^{2}},
    \label{eq:nig_pdf}
\end{equation}
and $K_{1}$ is the modified Bessel function of the second kind of order 1.

In general, the integral in~\eqref{eq:nig_cdf} admits no closed-form expression, and the CDF must therefore be computed numerically. In practice, we use the \texttt{norminvgauss} class from \texttt{scipy.stats} to compute reference values and assess surrogates' accuracy.

The training samples $\big(\hat{\boldsymbol{\theta}}^{(j)}, \hat{y}^{(j)}, \nabla_{\!\hat{\boldsymbol{\theta}}}\hat{y}^{(j)}\big)$, $j = 1,\ldots,J$, are generated according to
\begin{equation}
    \hat{y}^{(j)} = (b^{(j)} - a)\,
    f_{\mathrm{NIG}}\!\big(x^{(j)};\boldsymbol{\theta}^{(j)}\big),
    \qquad
    x^{(j)} = a + (b^{(j)} - a)\,u^{(j)}, \quad
    u^{(j)} \sim \mathcal{U}(0,1),
    \label{eq:nig_training_label}
\end{equation}
which provides unbiased single-realization MC labels of the target mapping. Here, the corresponding parameter differentials are
\begin{align*}
    \frac{\partial \hat{y}^{(j)}}{\partial b}
        &= f_{\mathrm{NIG}}\!\big(x^{(j)};\boldsymbol{\theta}^{(j)}\big)
        + (b^{(j)} - a)\,
          \frac{\partial}{\partial x}
          f_{\mathrm{NIG}}\!\big(x^{(j)};\boldsymbol{\theta}^{(j)}\big)\,
          u^{(j)}, \\[3pt]
    \nabla_{\!\boldsymbol{\theta}} \hat{y}^{(j)}
        &= (b^{(j)} - a)\,
           \nabla_{\!\boldsymbol{\theta}}
           f_{\mathrm{NIG}}\!\big(x^{(j)};\boldsymbol{\theta}^{(j)}\big),
\end{align*}
where the partial derivatives of the NIG PDF~\eqref{eq:nig_pdf} are given by
\begin{align*}
    \frac{\partial}{\partial x}
        f_{\mathrm{NIG}}\!\big(x^{(j)};\boldsymbol{\theta}^{(j)}\big)
        &= \beta\,f_{\mathrm{NIG}}\!\big(x^{(j)};\boldsymbol{\theta}^{(j)}\big)
         + K^{(j)} (x^{(j)} - \mu)
        = -\frac{\partial}{\partial \mu}
          f_{\mathrm{NIG}}\!\big(x^{(j)};\boldsymbol{\theta}^{(j)}\big), \\[3pt]
    \frac{\partial}{\partial \alpha}
        f_{\mathrm{NIG}}\!\big(x^{(j)};\boldsymbol{\theta}^{(j)}\big)
        &= f_{\mathrm{NIG}}\!\big(x^{(j)};\boldsymbol{\theta}^{(j)}\big)
           \left(
               \frac{1}{\alpha}
               + \frac{\delta \alpha}{\tau}
               + \frac{\upsilon^{(j)} K'_1(\alpha \upsilon^{(j)})}{K_1(\alpha \upsilon^{(j)})}
           \right), \\[3pt]
    \frac{\partial}{\partial \beta}
        f_{\mathrm{NIG}}\!\big(x^{(j)};\boldsymbol{\theta}^{(j)}\big)
        &= f_{\mathrm{NIG}}\!\big(x^{(j)};\boldsymbol{\theta}^{(j)}\big)
           \left(
               -\frac{\delta \beta}{\tau}
               + x^{(j)} - \mu
           \right), \\[3pt]
    \frac{\partial}{\partial \delta}
        f_{\mathrm{NIG}}\!\big(x^{(j)};\boldsymbol{\theta}^{(j)}\big)
        &= f_{\mathrm{NIG}}\!\big(x^{(j)};\boldsymbol{\theta}^{(j)}\big)
           \left(
               \frac{1}{\delta} + \tau
           \right)
           + \delta\,K^{(j)},
\end{align*}
with
\begin{equation*}
    K^{(j)} =
    \frac{\alpha \delta}{\pi [\upsilon^{(j)}]^{3}}
    \exp\!\big(\delta \tau + \beta (x^{(j)} - \mu)\big)
    \Big[
        \alpha \upsilon^{(j)} K'_1(\alpha \upsilon^{(j)}) - K_1(\alpha \upsilon^{(j)})
    \Big].
\end{equation*}

\begin{figure}[t!]
\centering
\begin{subfigure}[t]{0.48\textwidth}
    \centering
    \includegraphics[width=\linewidth]{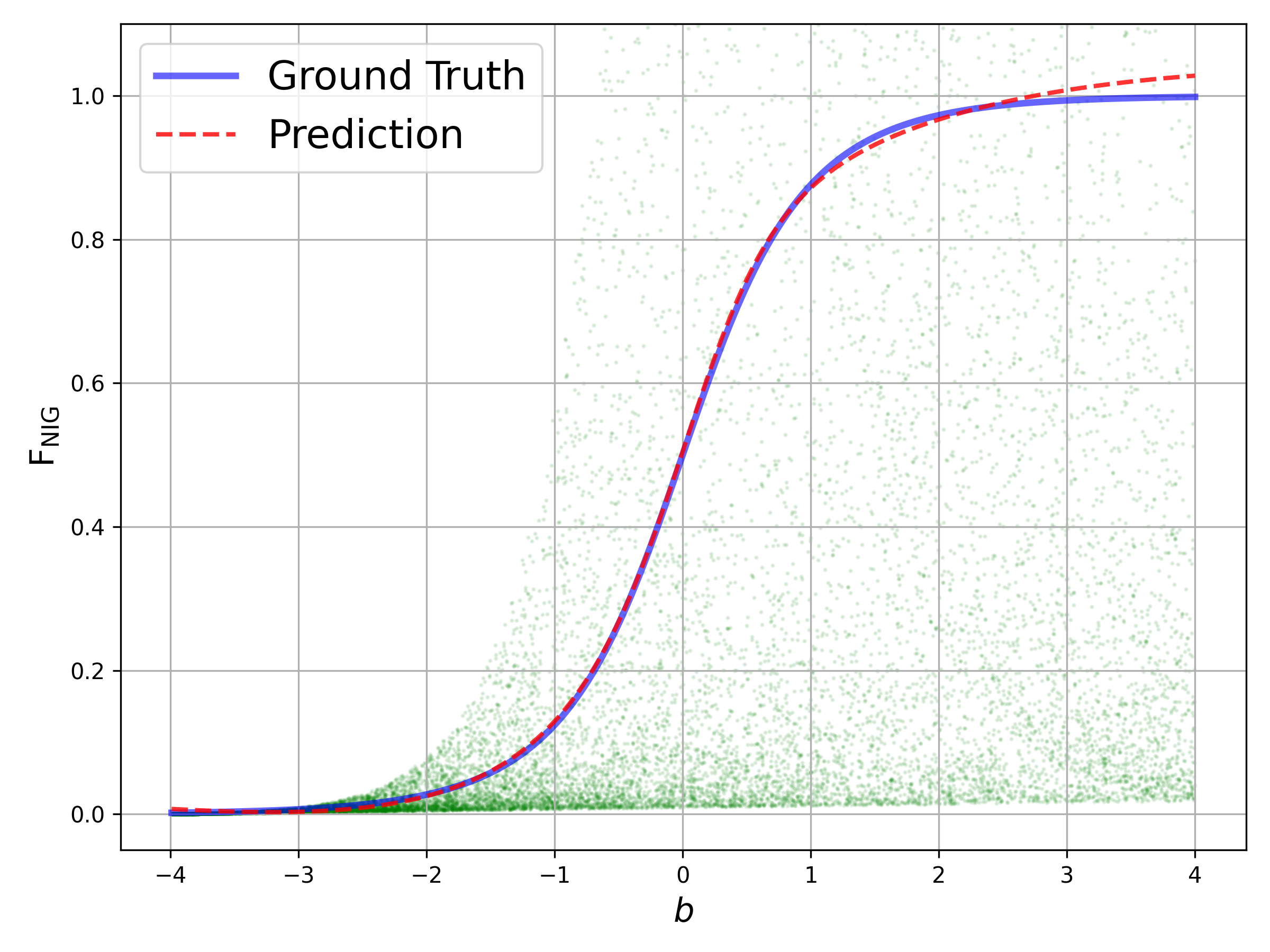}
\end{subfigure}
\hfill
\begin{subfigure}[t]{0.48\textwidth}
    \centering
    \includegraphics[width=\linewidth]{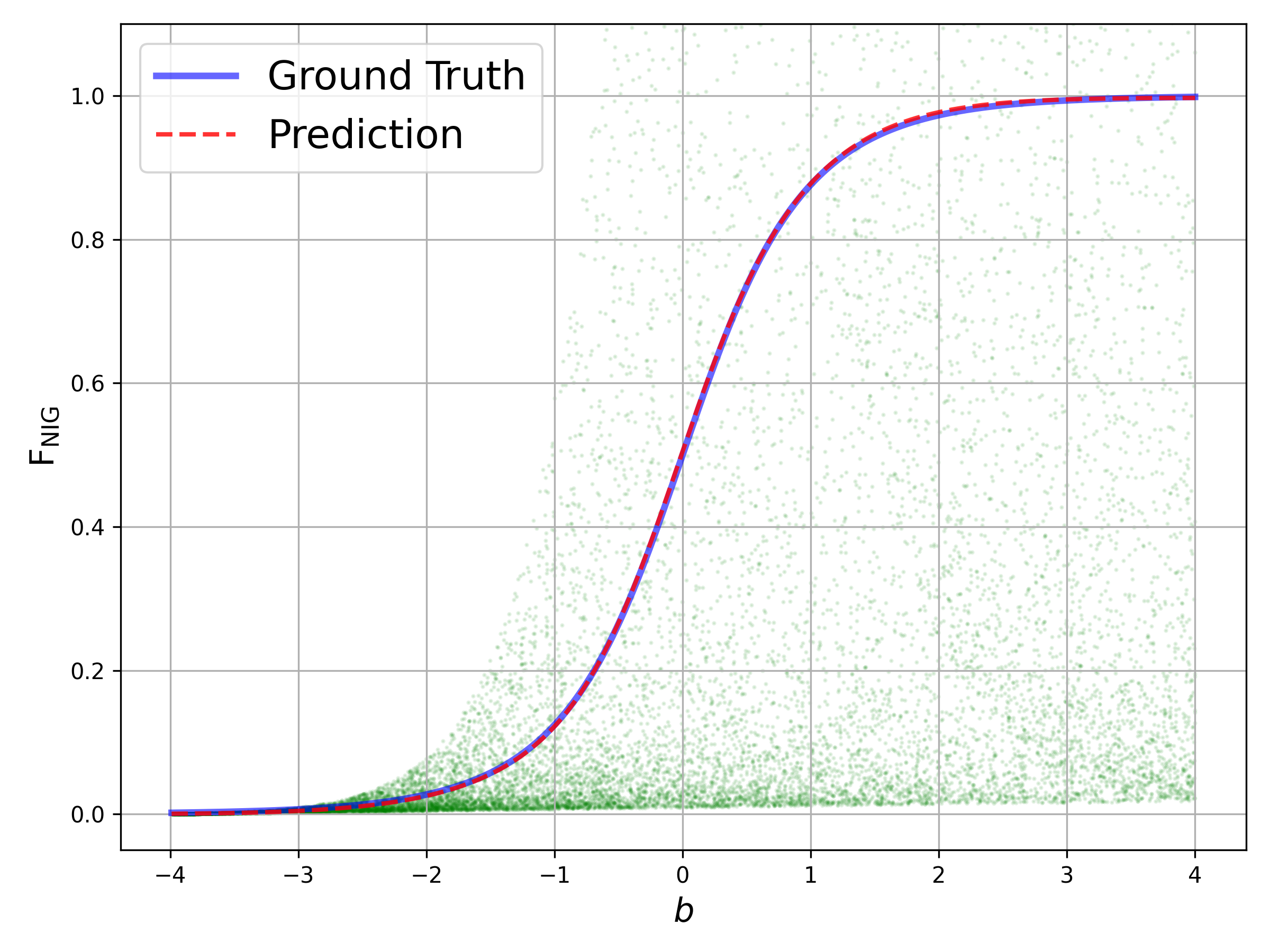}
\end{subfigure}
\caption{Predictions vs.\ analytical CDFs for ANN (left) and DML (right) for $X \sim \mathrm{NIG}(1,0,0,1)$, $a=-4$ and $b \in [-3.99,4]$.}
\label{fig:NIG_CDFs}
\end{figure}

\begin{figure}[t!]
\centering
\begin{subfigure}[t]{0.48\textwidth}
    \centering
    \includegraphics[width=\linewidth]{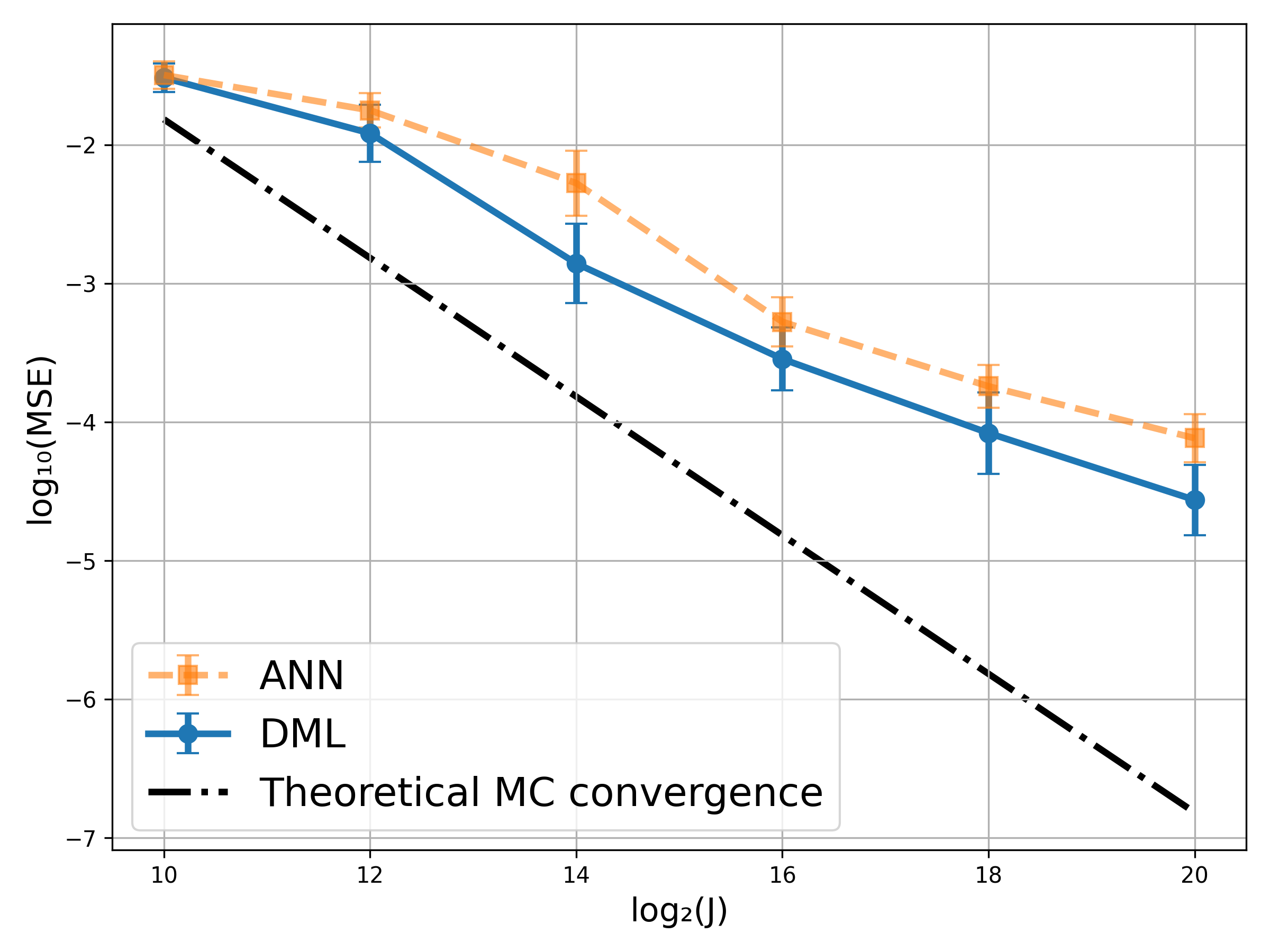}
    \caption*{$X \sim \chi^2(1)$, $b \in [0.01,10]$}
\end{subfigure}
\hfill
\begin{subfigure}[t]{0.48\textwidth}
    \centering
    \includegraphics[width=\linewidth]{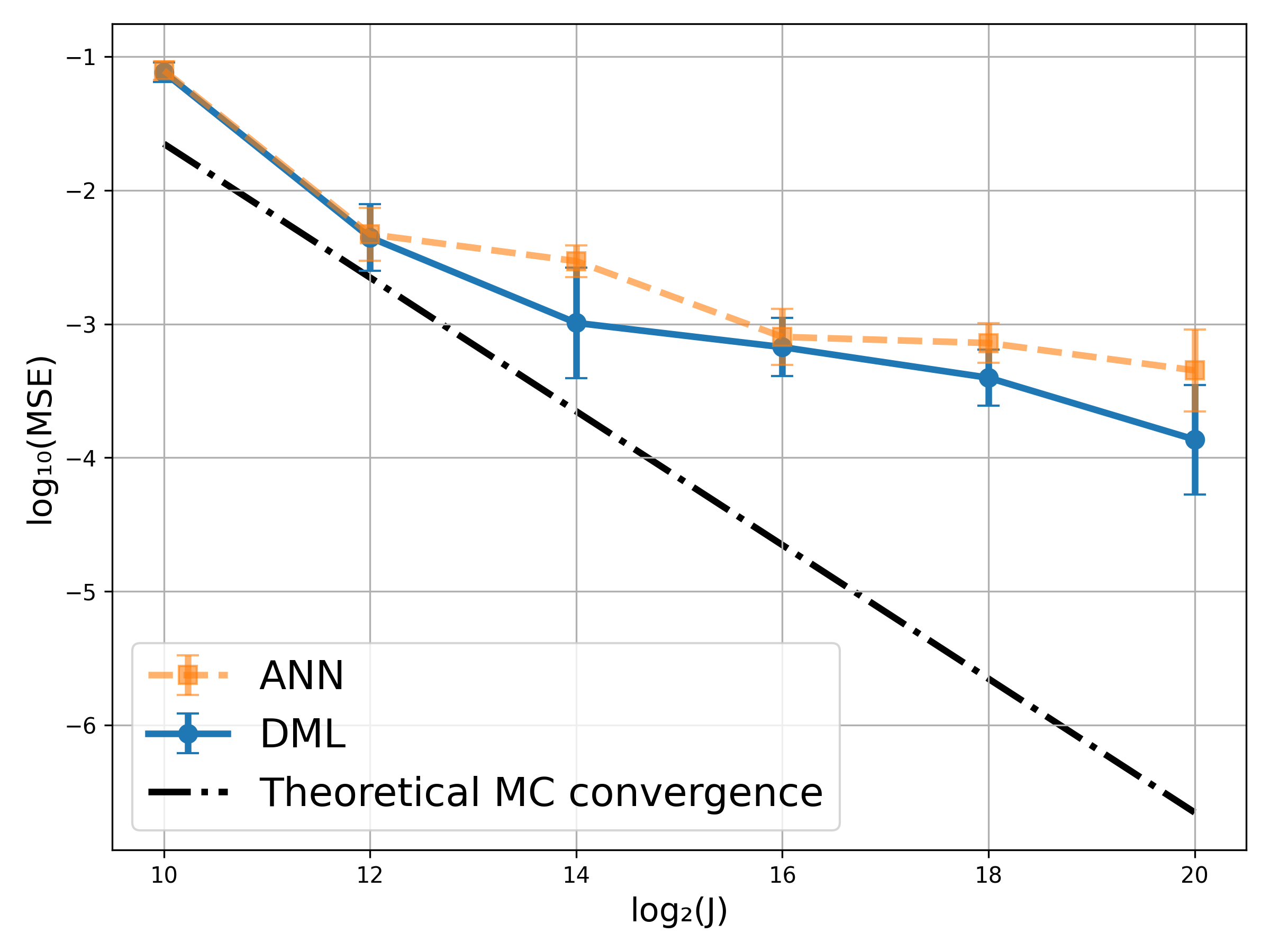}
    \caption*{$X \sim \chi^2(\theta)$, $b \in [0.01,10]$, $\theta \in [0.5,5]$}
\end{subfigure}
\vspace{0.3em}
\begin{subfigure}[t]{0.48\textwidth}
    \centering
    \includegraphics[width=\linewidth]{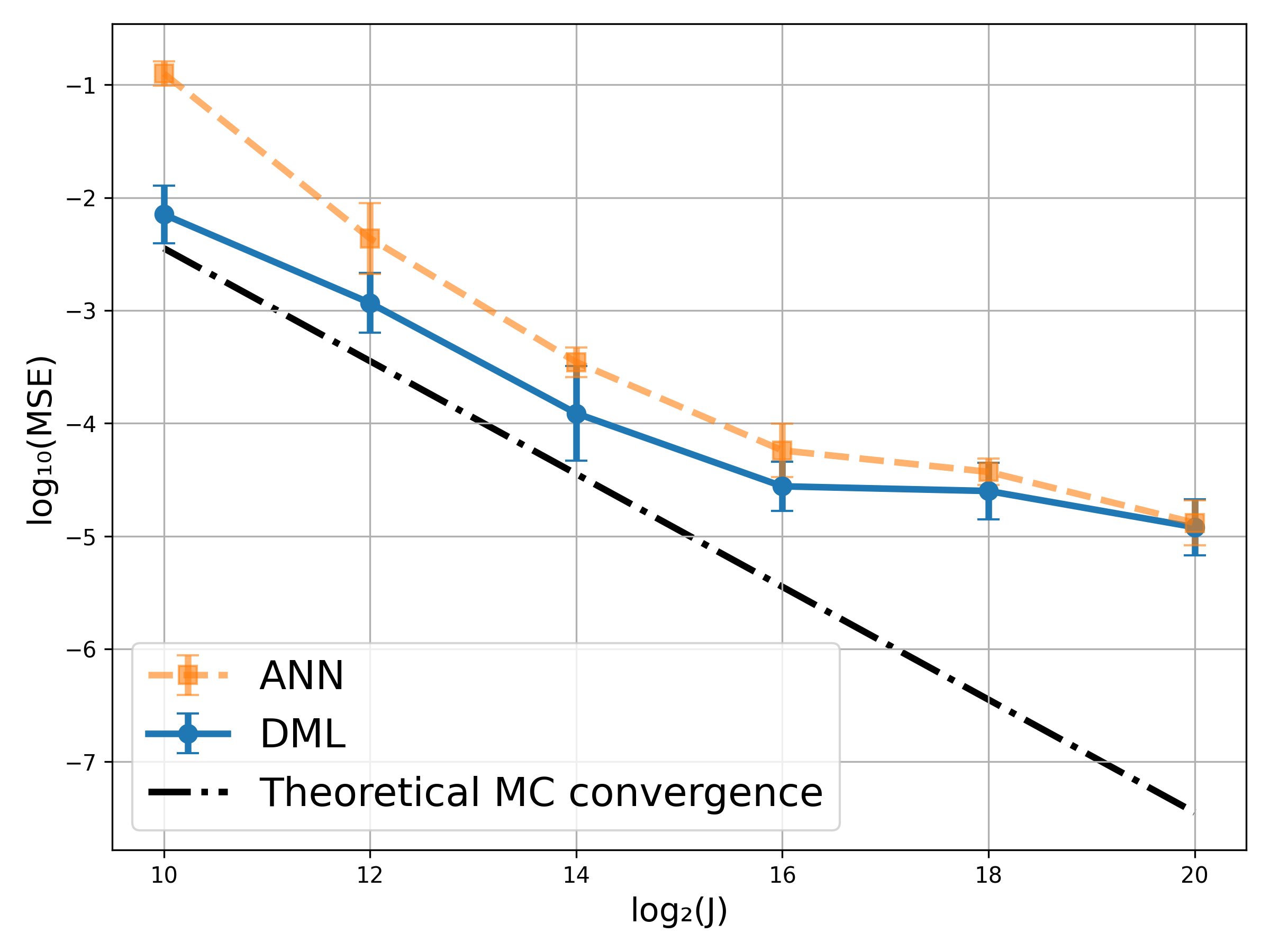}
    \caption*{$X \sim \mathrm{NIG}(1,0,0,1)$, $a = -4$, $b \in [-3.99,4]$}
\end{subfigure}
\hfill
\begin{subfigure}[t]{0.48\textwidth}
    \centering
    \includegraphics[width=\linewidth]{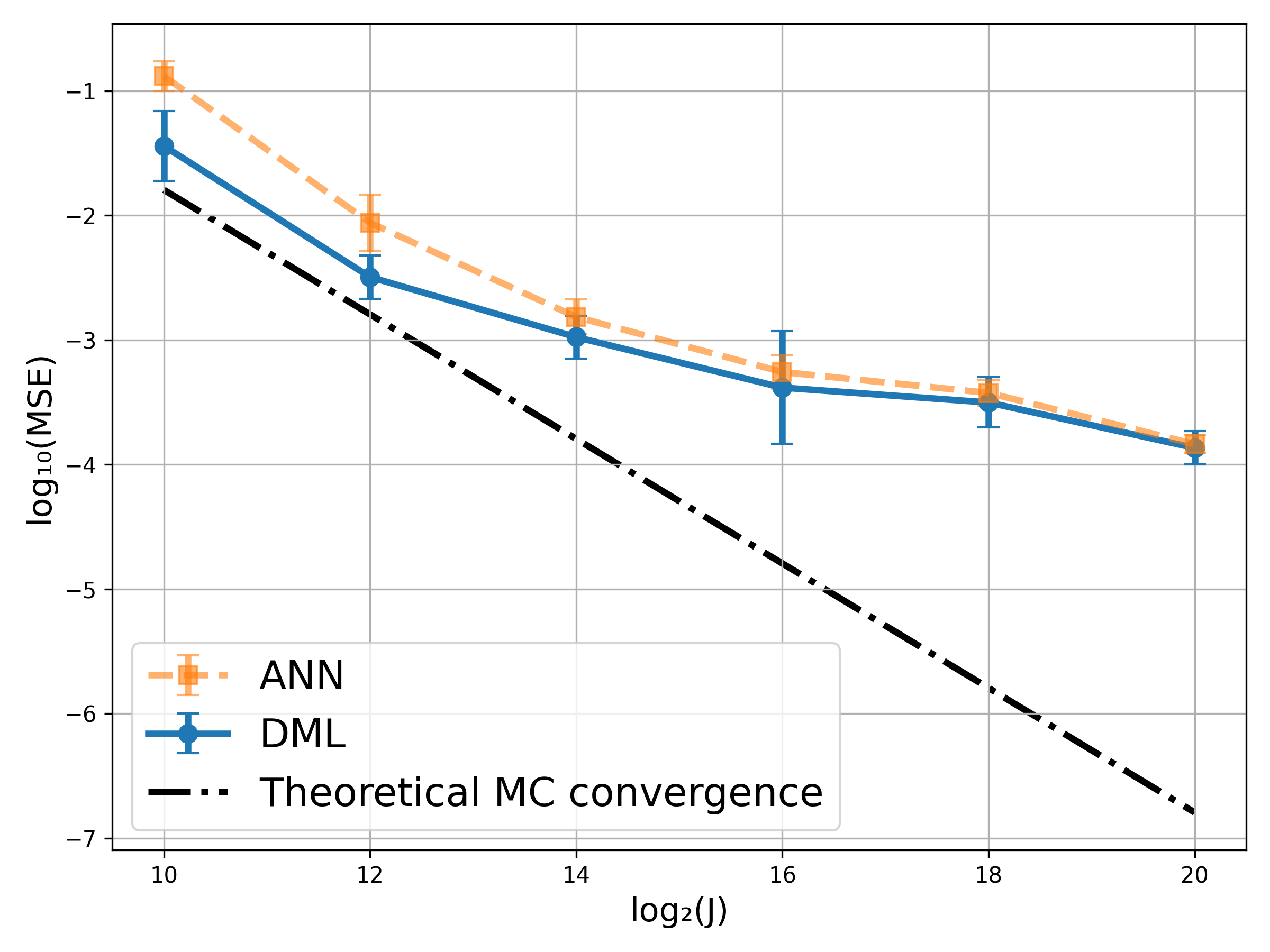}
    \caption*{$X \sim \mathrm{NIG}(\boldsymbol{\theta})$, $a = -4$, $b \in [-3.99,4]$, $\alpha \in [0.75,1]$, $\beta \in [-0.25,0.25]$, $\mu \in [-0.25,0.25]$, $\delta \in [0.75,1]$}
\end{subfigure}
\caption{MSE vs.\ training set size for parametric CDFs.}
\label{fig:CDFs}
\end{figure}

In Figure~\ref{fig:NIG_CDFs}, we present the results obtained for the benchmark examples of parametric CDFs under the NIG family for $J = 2^{16}$, comparing the predictions of the baseline ANN and the proposed DML framework for the one-input case $X \sim \mathrm{NIG}(1,0,0,1)$ with integration limits $a=-4$ and $b \in [-3.99,4]$. Figure~\ref{fig:CDFs} then summarizes the corresponding convergence behaviour, reporting the MSE as a function of the training set size for both the previously considered chi-squared and NIG examples, as well as for the five-input case $X \sim \mathrm{NIG}(\boldsymbol{\theta})$ with $a=-4$, $b \in [-3.99,4]$, $\alpha, \delta \in [0.75,1]$, and $\beta, \mu \in [-0.25,0.25]$.

\subsubsection{Discussion}

Across all experiments in Section~\ref{sec:param_statistics}, several consistent patterns emerge:
\begin{itemize}
    \item \textbf{Accuracy.} The DML framework consistently reproduces analytical and numerical targets with higher precision across all parameter domains. Figures~\ref{fig:lognormal_moments}, \ref{fig:lognormal_moments_2D}, \ref{fig:chi_CDFs} and~\ref{fig:chi_CDFs_2D} show that DML tracks the true curves almost exactly, even in regions of steep gradient or near parameter boundaries, while standard ANNs show visible inaccuracies. 
    \item \textbf{Sample Efficiency.} As evidenced in Figures~\ref{fig:moments} and~\ref{fig:CDFs}, the MSE of DML decays significantly faster with respect to dataset size $J$. DML achieves up to an order-of-magnitude reduction in MSE. This confirms the theoretical expectation that including differential information acts as a strong variance-reduction mechanism.
    \item \textbf{Scalability.} The benefits of DML persist across increasingly complex distributions, from univariate $\chi^2$ to four-parameter NIG families, demonstrating that gradient-enhanced training scales gracefully with the dimensionality of~$\boldsymbol{\theta}$. The figures illustrate that prediction quality remains uniform even when the number of inputs quintuples.
\end{itemize}

\subsection{Approximation of Parametric Functions}
\label{sec:param_functions}

We next investigate the approximation of parametric functions through orthogonal polynomial
expansions, with a specific focus on Chebyshev coefficients. Our experiments evaluate
the capability of both standard ANNs and DML models to learn multi-output mappings. 

\subsubsection{Parametric Chebyshev Coefficients}

Given a parametric function $f(x;\boldsymbol{\theta})$, its Chebyshev expansion on the interval $[-1,1]$ can be expressed as
\begin{equation}
    f(x;\boldsymbol{\theta})
    \approx \sum_{l=0}^{+\infty}{'} c_l(\boldsymbol{\theta})\,T_l(x),
    \qquad
    c_l(\boldsymbol{\theta})
    = \frac{2}{\pi} \int_{-1}^{1}
      \frac{f(x;\boldsymbol{\theta})\,T_l(x)}{\sqrt{1 - x^2}}\,
      \mathrm{d}x,
    \label{eq:chebyshev_coeff}
\end{equation}
where $T_l$ denotes the $l$-th Chebyshev polynomial and the notation $\sum\nolimits'$ indicates that the first term ($l=0$) is multiplied by $1/2$. Here, our goal is to approximate the multi-output mapping $\boldsymbol{\theta} \mapsto (c_0(\boldsymbol{\theta}),\ldots,c_L(\boldsymbol{\theta}))$, representing the first $L+1$ coefficients of the expansion.

\subsubsection*{Experiment 1: Exponential Function}

We begin by considering the exponential function $f(x;\theta) = \exp(\theta x)$. In this case, the Chebyshev coefficients admit analytical expressions,
\begin{equation*}
    c_0(\theta) = I_0(\theta), \qquad c_l(\theta) = 2 I_l(\theta), \quad l \ge 1,
\end{equation*}
where $I_l$ denotes the modified Bessel function of the first kind of order $l$. These closed-form results provide an exact reference to assess model accuracy.

Training data are generated by uniform sampling in $x \in [-1,1]$. For each parameter sample $\theta^{(j)}$ and random $x^{(j)} = -1 + 2 u^{(j)}$ with $u^{(j)} \sim \mathcal{U}(0,1)$, the single-realization MC label and its differential are defined as
\begin{equation*}
    \hat{y}^{(j)}_l
    = \frac{4}{\pi}
      \frac{e^{\theta^{(j)} x^{(j)}}\,T_l(x^{(j)})}
           {\sqrt{1 - [x^{(j)}]^2}},
    \qquad
    \frac{\partial \hat{y}^{(j)}_l}{\partial \theta}
    = x^{(j)}\,\hat{y}^{(j)}_l,
\end{equation*}
for $j = 1,\ldots,J$ and $l = 1,\ldots,L$. 

\begin{figure}[t!]
\centering
\begin{subfigure}{0.48\textwidth}
  \centering
  \includegraphics[width=\linewidth]{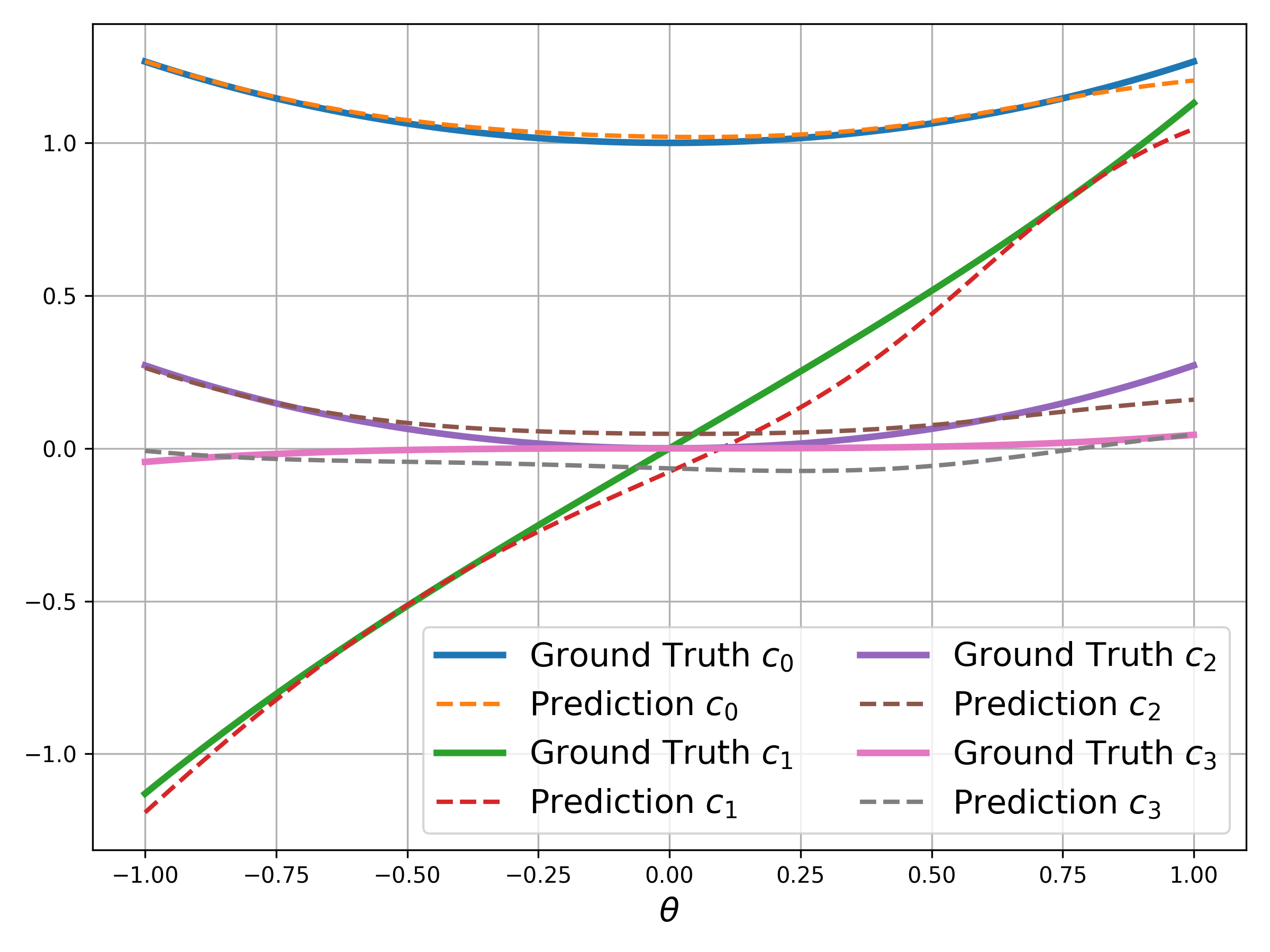}
\end{subfigure}
\hfill
\begin{subfigure}{0.48\textwidth}
  \centering
  \includegraphics[width=\linewidth]{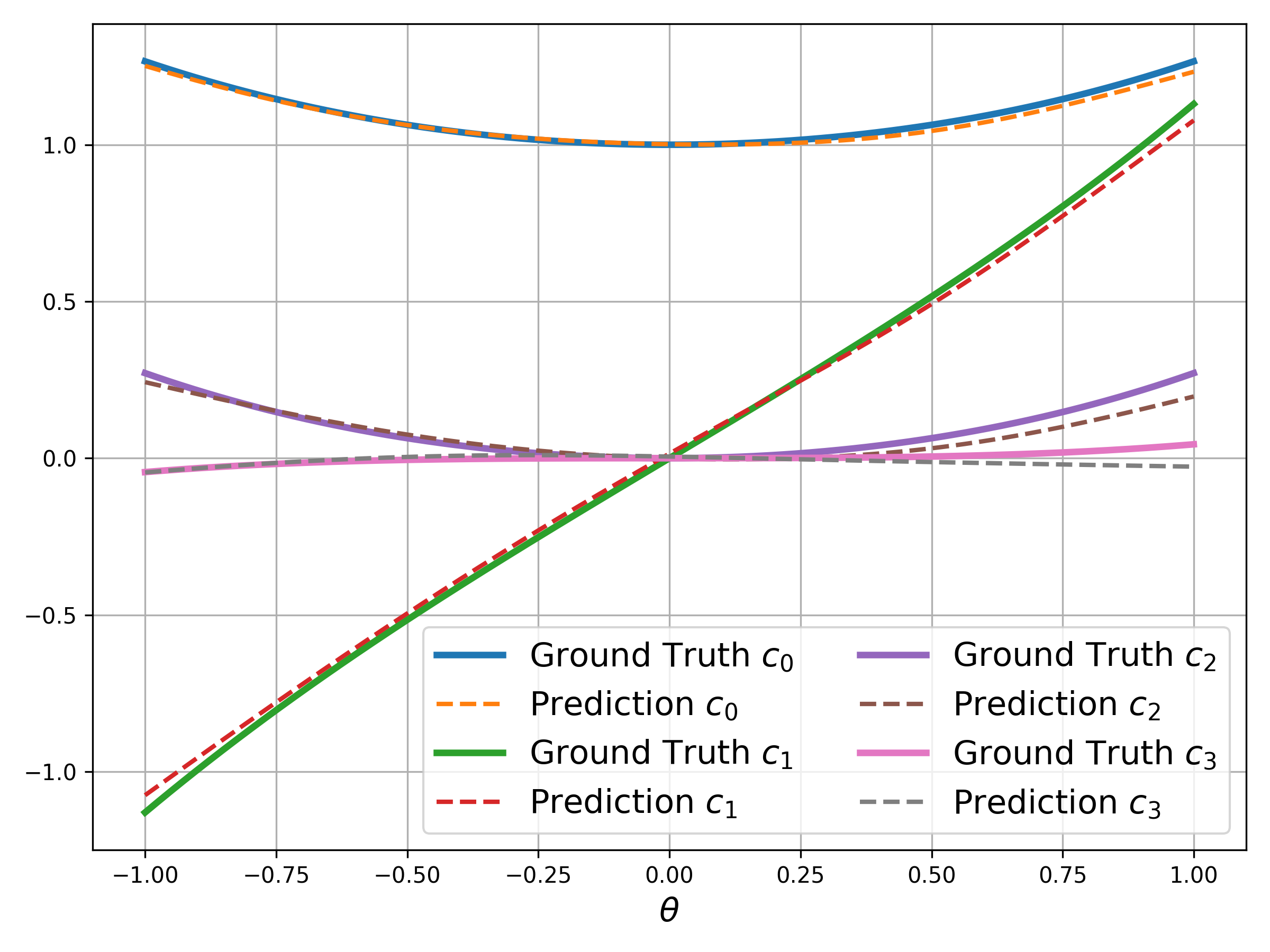}
\end{subfigure}
\caption{Predictions vs.\ analytical Chebyshev coefficients for ANN (left) and DML (right) for the exponential function with $\theta \in [-1,1]$ and $L = 15$.}
\label{fig:cheb_exp}
\end{figure}

Figure~\ref{fig:cheb_exp} compares the predictions of the baseline ANN and the proposed DML framework for $f(x;\theta) = \exp(\theta x)$ with $\theta \in [-1,1]$ and $L=15$, displaying the first four Chebyshev coefficients for $J = 2^{16}$.  

\subsubsection*{Experiment 2: Piecewise Exponential--Quadratic Function}

We next consider the piecewise-defined function
\begin{equation}
    f(x;\boldsymbol{\theta}) =
    \begin{cases}
        e^{\xi x}, & x \in [-1,0], \\[3pt]
        A x^{2} + B x + C, & x \in (0,1],
    \end{cases}
    \label{eq:piecewise_function}
\end{equation}
where $\boldsymbol{\theta} = (\xi, A, B, C)$. In this case, no analytical expressions for the Chebyshev coefficients exist, and they must therefore be computed numerically. Reference values are obtained using the Gauss–Chebyshev quadrature rule, which provides high-accuracy approximations for integrals of the form~\eqref{eq:chebyshev_coeff}.

The training dataset $\big(\boldsymbol{\theta}^{(j)}, \hat{y}^{(j)}_l, \nabla_{\!\boldsymbol{\theta}}\hat{y}^{(j)}_l\big)$, with $j = 1,\ldots,J$ and $l = 0,\ldots,L$, is generated as
\begin{equation*}
    \hat{y}^{(j)}_l
    = \frac{4}{\pi}\,
      \frac{f(x^{(j)};\boldsymbol{\theta}^{(j)})\,T_l(x^{(j)})}
           {\sqrt{1 - [x^{(j)}]^2}},
    \qquad
    x^{(j)} = -1 + 2u^{(j)}, \quad u^{(j)} \sim \mathcal{U}(0,1),
\end{equation*}
providing unbiased single-realization MC labels of the target coefficients. The parameter differentials used in the loss \eqref{eq:combined_loss} are given by
\begin{align*}
    \frac{\partial \hat{y}^{(j)}_l}{\partial \xi}
        &= \mathbf{1}_{\{x^{(j)} \le 0\}}\,
           x^{(j)}\,\hat{y}^{(j)}_l, &
    \frac{\partial \hat{y}^{(j)}_l}{\partial A}
        &= \mathbf{1}_{\{x^{(j)} > 0\}}\,
           [x^{(j)}]^2\,\hat{y}^{(j)}_l, \nonumber \\[3pt]
    \frac{\partial \hat{y}^{(j)}_l}{\partial B}
        &= \mathbf{1}_{\{x^{(j)} > 0\}}\,
           x^{(j)}\,\hat{y}^{(j)}_l, &
    \frac{\partial \hat{y}^{(j)}_l}{\partial C}
        &= \mathbf{1}_{\{x^{(j)} > 0\}}\,
           \hat{y}^{(j)}_l.
\end{align*} 

\begin{figure}[t!]
\centering
\begin{subfigure}{0.48\textwidth}
  \centering
  \includegraphics[width=\linewidth]{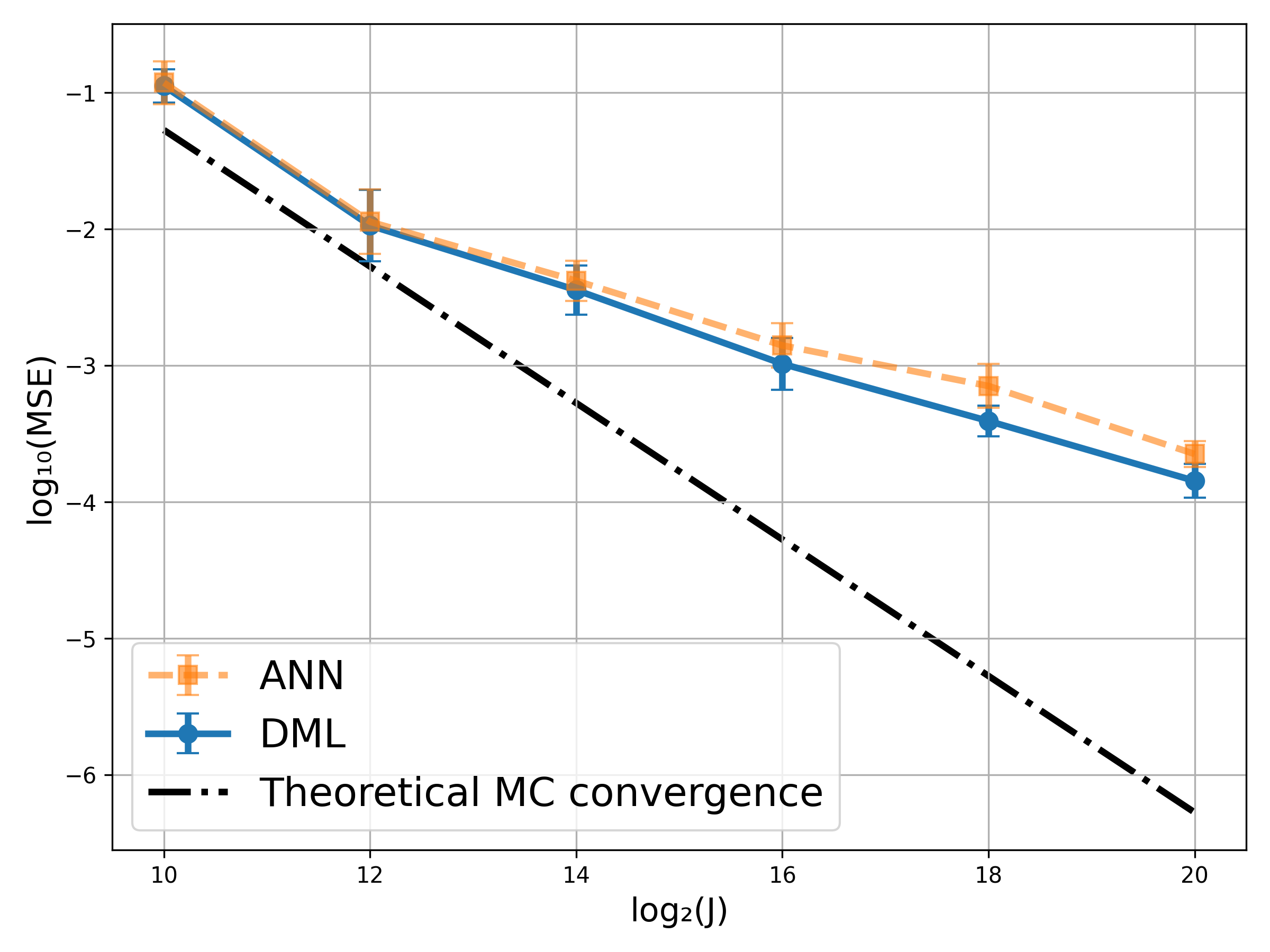}
\end{subfigure}
\hfill
\begin{subfigure}{0.48\textwidth}
  \centering
  \includegraphics[width=\linewidth]{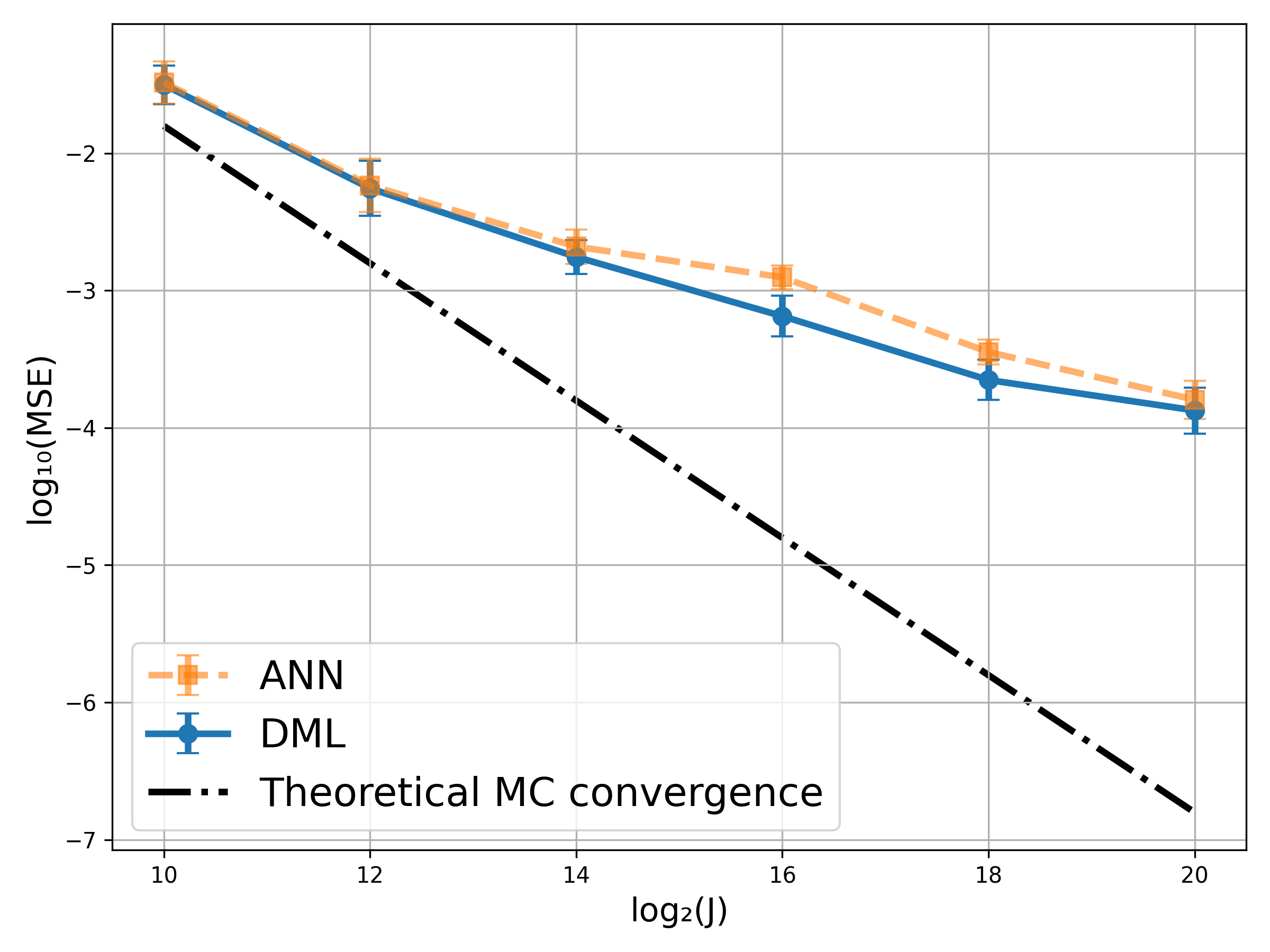}
\end{subfigure}
\vspace{0.3em}
\begin{subfigure}{0.48\textwidth}
  \centering
  \includegraphics[width=\linewidth]{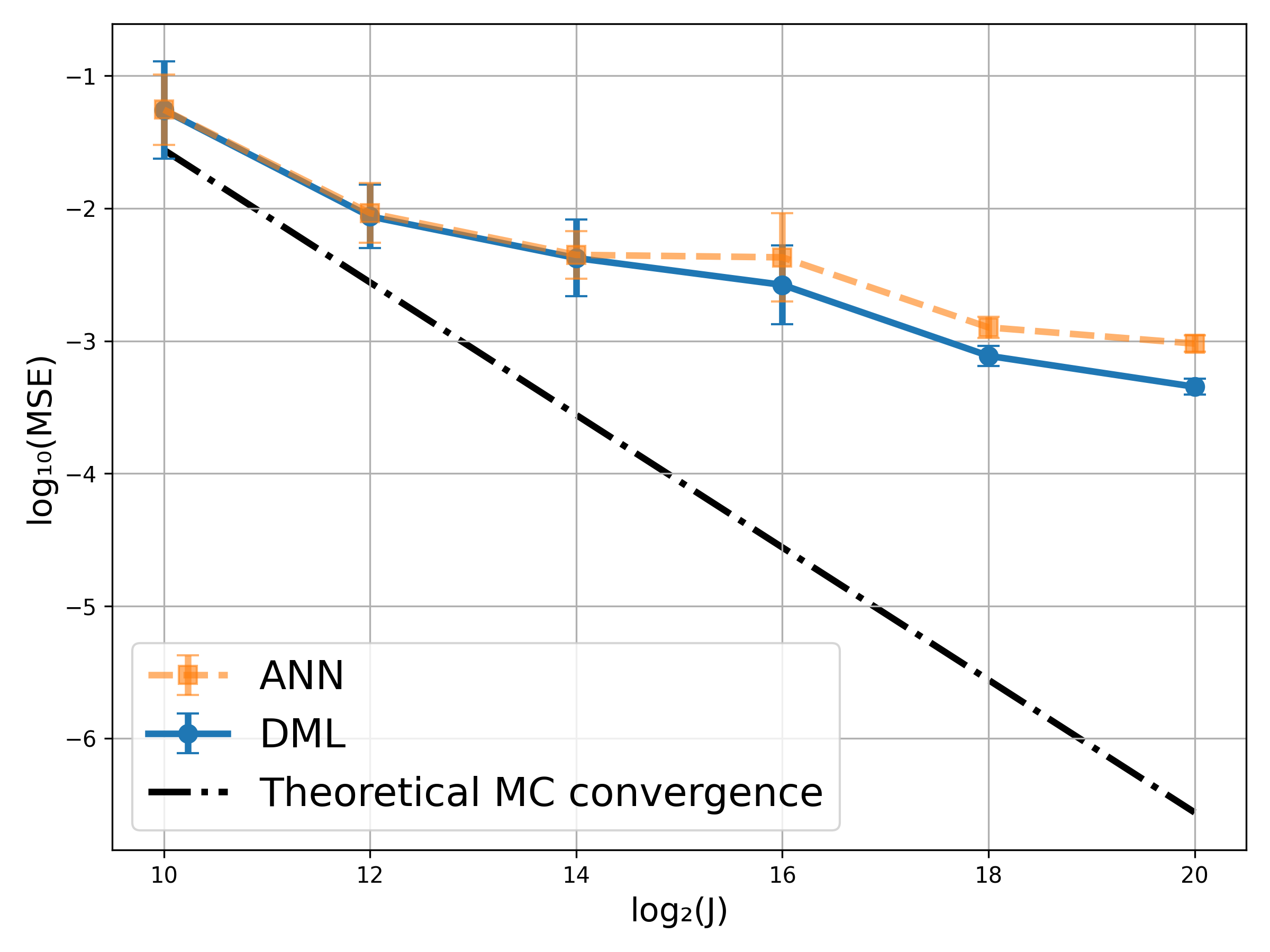}
\end{subfigure}
\hfill
\begin{subfigure}{0.48\textwidth}
  \centering
  \includegraphics[width=\linewidth]{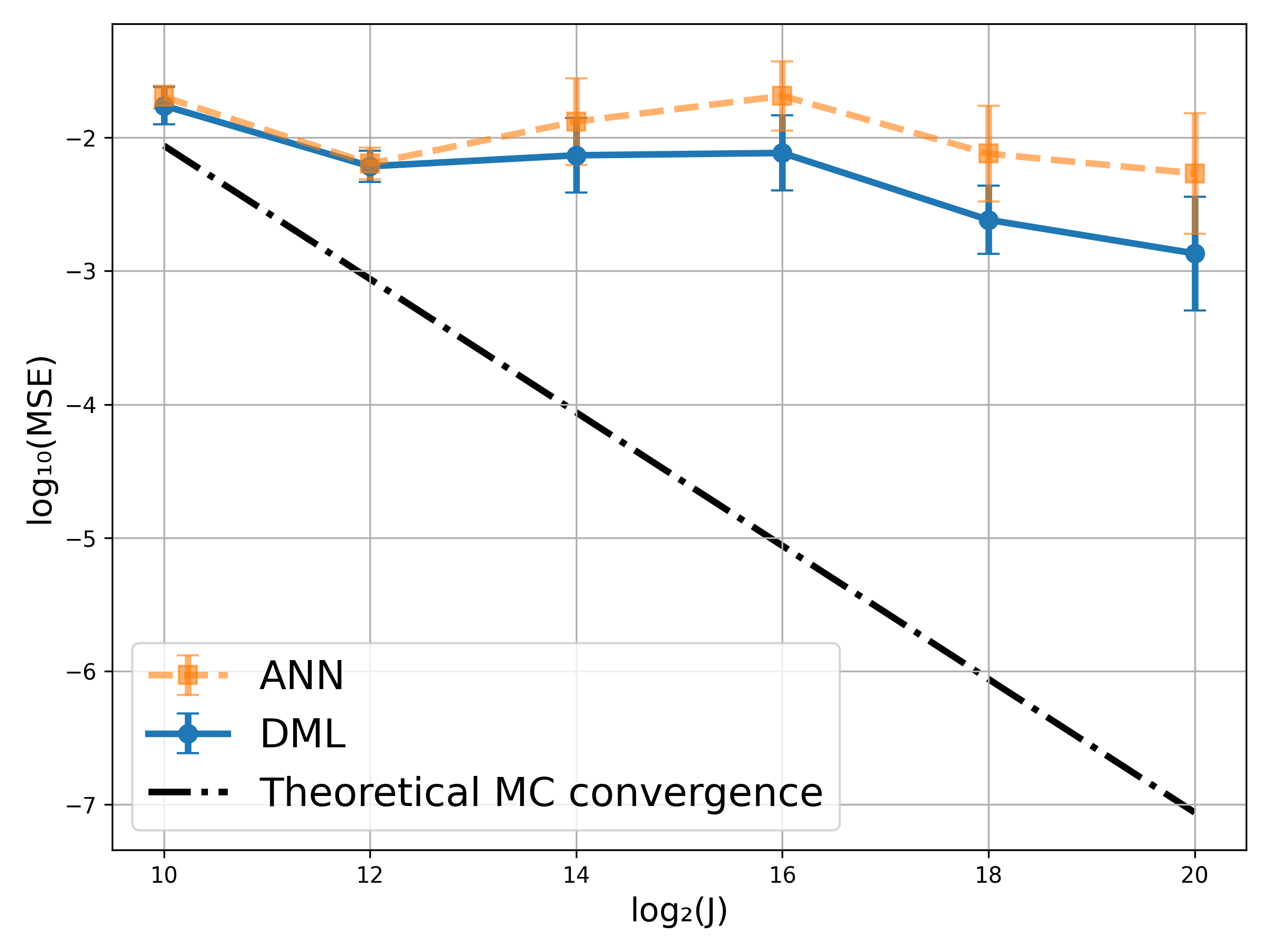}
\end{subfigure}
\caption{Cumulative MSE vs.\ training set size for $L = 1$ (left column) and $L = 15$ (right column), for the exponential function (top row) with $\theta \in [-1,1]$, and the piecewise exponential–quadratic function (bottom row) with $\xi \in [0.1, 2]$ and $A, B, C \in [-1, 1]$.}
\label{fig:Chebyshev_coefficients}
\end{figure}

In Figure~\ref{fig:Chebyshev_coefficients}, we summarize the convergence behaviour, reporting the cumulative MSE as a function of the training set size for $f(x;\theta) = \exp(\theta x)$, $\theta \in [-1,1]$ and the piecewise exponential–quadratic function \eqref{eq:piecewise_function}, $\xi \in [0.1, 2]$, $A, B, C \in [-1, 1]$, each evaluated for $L=1$ and $L=15$.

\subsubsection{Discussion}

The numerical experiments in Section~\ref{sec:param_functions} reveal several consistent and recurrent patterns:
\begin{itemize}
    \item \textbf{Accuracy.} For both smooth (exponential) and non-smooth (piecewise exponential--quadratic) functions, DML always attains a better estimation of the Chebyshev coefficients than basic ANNs. Figure~\ref{fig:cheb_exp} shows that DML more accurately approximates the true coefficients in regions of high curvature or near parameter boundaries compared with ANN models. 
    \item \textbf{Sample Efficiency.} From Figure~\ref{fig:Chebyshev_coefficients}, DML framework reaches the same error level as the standard ANN using fewer training samples, yielding a reduction in cumulative MSE of more than half an order of magnitude. Further, for case of the piecewise exponential-quadratic case, the convergence of standard ANN approaches even diverges around $J=2^{14}$.
    \item \textbf{Scalability.} The differential formulation naturally extends to high-dimensional parameter domains and multi-output mappings, with DML showing improved scalability as the number of outputs grows from two to sixteen.
\end{itemize}

\subsection{Integrals derived from Differential Equations} \label{sec:param_ODEs_PDEs}

Beyond statistical integrals and function-approximation settings, we now consider integrals that naturally arise from ODEs and PIDEs. Such integrals are ubiquitous in, for example, physics, engineering, or quantitative finance. 

\subsubsection{Parametric Integrals derived from ODEs}
\label{sec:ODEintegrals}

We begin by analysing parametric integrals that arise directly from ODEs. A prototypical example is provided by the nonlinear pendulum, whose solution involves an incomplete elliptic integral.

\subsubsection*{Experiment 1: The Nonlinear Pendulum}

We begin considering the motion of a simple pendulum governed by the nonlinear ODE
\begin{equation}
\frac{\mathrm{d}^2\phi}{\mathrm{d}t^2} + \frac{g}{R}\sin(\phi) = 0,
\label{eq:nonlinear_pendulum}
\end{equation}
where $\phi(t)$ denotes the angular displacement at time $t$, $g$ is the gravitational acceleration, and $R$ is the pendulum length. Equation~\eqref{eq:nonlinear_pendulum} admits an energy integral of motion,
\[
\frac{1}{2}\Big(\frac{\mathrm{d}\phi}{\mathrm{d}t}\Big)^2
= \frac{g}{R}\big(\cos(\phi) - \cos(\phi_0)\big),
\]
where $\phi_0$ is the maximum angular amplitude. By separating variables, the time--angle relation is obtained, which gives the time required for the pendulum to swing from the vertical position to a given angle~$\phi$:
\begin{equation}
t(\phi;\phi_0) 
= \sqrt{\frac{R}{2g}}
\int_{0}^{\phi} 
\frac{\mathrm{d}\varphi}{\sqrt{\cos(\varphi) - \cos(\phi_0)}}.
\label{eq:time_angle_relation}
\end{equation}

Introducing the elliptic modulus $\theta = \sin(\phi_0/2)$ and the substitution $\sin(\varphi/2) = \theta \sin(\phi)$, the integral~\eqref{eq:time_angle_relation} reduces to
\begin{equation*}
t(\phi;\phi_0)
= 2\sqrt{\frac{R}{g}}\,
F\!\left(b; \theta\right),
\qquad
b = \arcsin\!\left(\frac{\sin(\phi/2)}{\theta}\right),
\end{equation*}
where 
\begin{equation}
\mathrm{F}(b; \theta) = \int_{0}^{b} 
\frac{\mathrm{d}x}{\sqrt{1 - \theta^2\sin^2(x)}},
\label{eq:incomplete_elliptic}
\end{equation}
denotes the incomplete Legendre elliptic integral of the first kind.

In general, the integral in \eqref{eq:incomplete_elliptic} admits no closed-form expression, and it must therefore be computed numerically. In practice, we use the \texttt{ellipkinc} class from \texttt{scipy.stats} to compute reference values and asses surrogate accuracy.

Let $\hat{\boldsymbol{\theta}} = (b,\theta)$. The training samples $(\hat{\boldsymbol{\theta}}^{(j)},\hat{y}^{(j)},\nabla_{\!\hat{\boldsymbol{\theta}}}\hat{y}^{(j)})$, $j = 1,\ldots,J$, are constructed as single-realization MC labels
\begin{equation*}
\hat{y}^{(j)} = \frac{b^{(j)}}{\sqrt{1 - [\theta^{(j)}]^2\sin^2(x^{(j)})}}, \qquad x^{(j)} = b^{(j)} u^{(j)}, \quad u^{(j)} \sim \mathcal{U}(0,1),
\end{equation*}
which provides unbiased estimators of \eqref{eq:incomplete_elliptic}. The corresponding input differentials, required for the loss \eqref{eq:combined_loss}, are computed as
\begin{equation*}
\frac{\partial \hat{y}^{(j)}}{\partial b}
= \frac{1-[\theta^{(j)}]^2\sin^2 x^{(j)} + \tfrac{1}{2}b^{(j)}[\theta^{(j)}]^2 u^{(j)}\sin(2x^{(j)})}
{\sqrt{\big(1-[\theta^{(j)}]^2\sin^2(x^{(j)})\big)^3}}, \qquad \frac{\partial \hat{y}^{(j)}}{\partial \theta}
= \frac{b^{(j)}\theta^{(j)}\sin^2(x^{(j)})}{\sqrt{\big(1-[\theta^{(j)}]^2\sin^2(x^{(j)})\big)^3}}.
\end{equation*}

\begin{figure}[t!]
\centering
\includegraphics[width=0.5\textwidth]{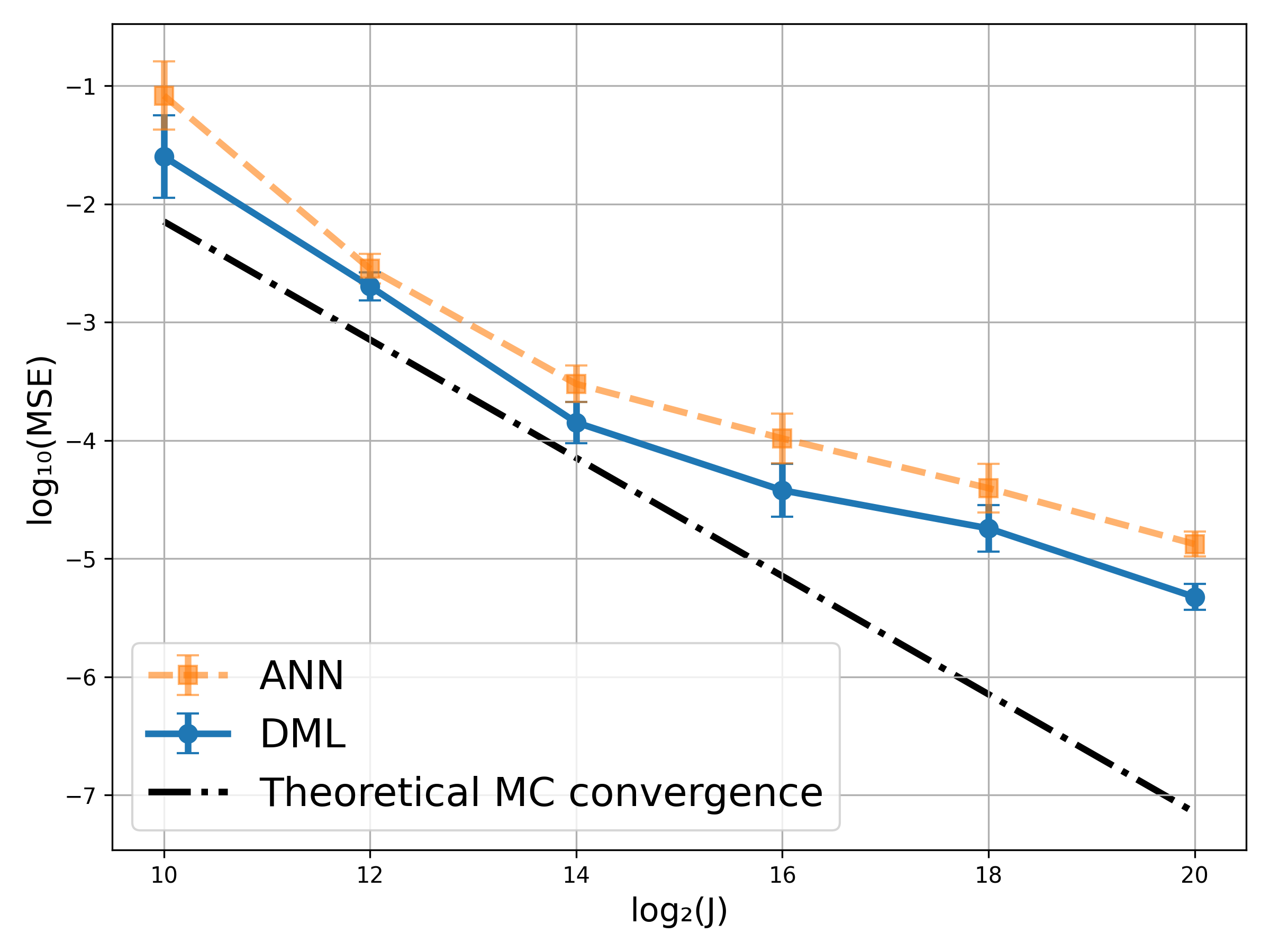}
\caption{MSE vs.\ training set size for the Legendre elliptic integral with \( b \in [0.01,\pi/2] \) and \( \theta \in [0,0.99] \).}
\label{fig:legendre}
\end{figure}

In Figure~\ref{fig:legendre}, we report the MSE as a function of the training set size for the incomplete Legendre elliptic integral with parameters \( b \in [0.01, \pi/2] \) and \( \theta \in [0, 0.99] \).

\subsubsection{Integrals derived from PIDEs}
\label{sec:PIDEintegrals}

We now analyse parametric integrals that naturally arise in PIDEs associated with stochastic processes exhibiting discontinuous jumps. As a representative case, we consider the Kou model~\cite{Kou2002}. 

\subsubsection*{Experiment 2: The Kou Model} 
\label{sec:kou_jump_diffusion}

Let us consider the one--dimensional asset price process $\{S_t\}_{t \ge 0}$, which evolves under the risk--neutral measure according to the stochastic differential equation
\begin{equation*}
\mathrm{d}S_t
= (r - \lambda \kappa) S_t\,\mathrm{d}t
+ \sigma_K S_t\,\mathrm{d}W_t
+ S_t \int_{\mathbb{R}} \big(e^{x}-1\big)\,N(\mathrm{d}t,\mathrm{d}x),
\end{equation*}
where \(r\) denotes the continuously compounded risk--free rate, \(\sigma_K>0\) is the Kou model's volatility, and \(W_t\) is a standard Brownian motion. The jump component is governed by a compensated Poisson random measure \(N(\mathrm{d}t,\mathrm{d}x)\) with constant intensity \(\lambda>0\), and jump size \(X\) distributed according to the double exponential density
\begin{equation*}
f_X(x; p, \eta_1, \eta_2)
= p\,\eta_1 e^{-\eta_1 x}\,\mathbf{1}_{\{x \ge 0\}}
+ (1-p)\,\eta_2 e^{\eta_2 x}\,\mathbf{1}_{\{x < 0\}},
\end{equation*}
where \(p \in [0,1]\) is the probability of an upward jump, and \(\eta_1, \eta_2 > 0\) are the decay rates of positive and negative jumps, respectively. The compensator term \(\kappa = \mathbb{E}[e^{X}-1]\) ensures risk neutrality.

Under suitable regularity conditions, the value function \(V(S,t)\) of a European--style derivative satisfies the backward PIDE
\begin{equation}
\frac{\partial V}{\partial t}(S,t) + \frac{1}{2}\sigma_K^2 S^2\,\frac{\partial^2 V}{\partial S^2}(S,t) + (r - \lambda \kappa)S\,\frac{\partial V}{\partial S}(S,t) - (r + \lambda)V(S,t) + \mathcal{J}[V](S,t) = 0,
\label{eq:kouPIDE}
\end{equation}
with nonlocal jump operator
\begin{equation}
\mathcal{J}[V](S,t)
= \lambda
\int_{\mathbb{R}}
\Big[
V\!\big(S e^{x},t\big) - S\,(e^{x}-1)\,\frac{\partial V}{\partial S}(S,t) \Big] f_X(x; p, \eta_1, \eta_2)\,\mathrm{d}x.
\label{eq:kou_jump_operator}
\end{equation}

To test the PIDE~\eqref{eq:kouPIDE}, we adopt a separable power--law ansatz of the form
\begin{equation}
V(S,t)=S^{2}\,e^{-\rho (T-t)},
\label{eq:kou_powerlaw_ansatz}
\end{equation}
where
\begin{equation*}
\rho = - r - \sigma_K^{2} + \lambda\Big(1 + 4\kappa - \mathbb{E}\!\big[e^{2X}\big]\Big).
\end{equation*}

Let $\boldsymbol{\theta} = (p,\eta_1,\eta_2)$. Using \eqref{eq:kou_powerlaw_ansatz}, the jump integral operator~\eqref{eq:kou_jump_operator} can be expressed compactly as
\begin{equation*}
\mathcal{J}[V](S,t)
= \lambda\,S^2\,e^{-\rho (T-t)}\mathrm{K}(\boldsymbol{\theta}),
\end{equation*}
where
\begin{equation} \label{eq:kou_Mtheta_integral}
\mathrm{K}(\boldsymbol{\theta})
= \int_{\mathbb{R}} \Big[e^{2x} - 2(e^{x}-1) \Big] f_X(x; p, \eta_1, \eta_2)\,\mathrm{d}x.
\end{equation}

The integral~\eqref{eq:kou_Mtheta_integral} admits the closed--form expression
\begin{equation*}
\mathrm{K}(\boldsymbol{\theta})
= p\,\frac{\eta_1}{\eta_1 - 2}
+ (1-p)\,\frac{\eta_2}{\eta_2 + 2}
- 2\!\left(
p\,\frac{\eta_1}{\eta_1 - 1}
+ (1-p)\,\frac{\eta_2}{\eta_2 + 1}
- 1
\right),
\end{equation*}
which provides an exact analytical reference for surrogate model validation.

The training dataset $(\boldsymbol{\theta}^{(j)},\hat{y}^{(j)},\nabla_{\!\boldsymbol{\theta}}\hat{y}^{(j)})$, $j = 1,\ldots,J$, is generated as
\begin{equation*}
\hat{y}^{(j)} 
= (b-a)\,\big[e^{2x^{(j)}} - 2(e^{x^{(j)}}-1)\big] f_X(x^{(j)};\boldsymbol{\theta}^{(j)}),
\qquad
x^{(j)} = a + (b-a)\,u^{(j)}, \quad u^{(j)} \sim \mathcal{U}(0,1)
\end{equation*}
which yields an unbiased MC estimator of \eqref{eq:kou_Mtheta_integral}. The corresponding parameter differentials, required for the gradient-based loss~\eqref{eq:combined_loss}, are obtained from
\begin{equation*}
\nabla_{\!\boldsymbol{\theta}} \hat{y}^{(j)} 
= (b-a)\,\big[e^{2x^{(j)}} - 2(e^{x^{(j)}}-1)\big] \nabla_{\!\boldsymbol{\theta}} f_X(x^{(j)};\boldsymbol{\theta}^{(j)}),
\end{equation*}
where the partial derivatives of the PDF are
\begin{align*}
\frac{\partial f_X}{\partial p} 
&= \mathbf{1}_{\{x^{(j)} \ge 0\}}\,\eta_1^{(j)} e^{-\eta_1^{(j)} x^{(j)}}
 - \mathbf{1}_{\{x^{(j)} < 0\}}\,\eta_2^{(j)} e^{\eta_2^{(j)} x^{(j)}}, \\[1em]
\frac{\partial f_X}{\partial \eta_1} 
&= \mathbf{1}_{\{x^{(j)} \ge 0\}}\,p^{(j)} e^{-\eta_1^{(j)} x^{(j)}} \big(1 - \eta_1^{(j)} x^{(j)}\big), \\[1em]
\frac{\partial f_X}{\partial \eta_2} 
&= \mathbf{1}_{\{x^{(j)} < 0\}}\,(1-p^{(j)}) e^{\eta_2^{(j)} x^{(j)}} \big(1 + \eta_2^{(j)} x^{(j)}\big).
\end{align*}

\begin{figure}[t!]
\centering
\includegraphics[width=0.5\textwidth]{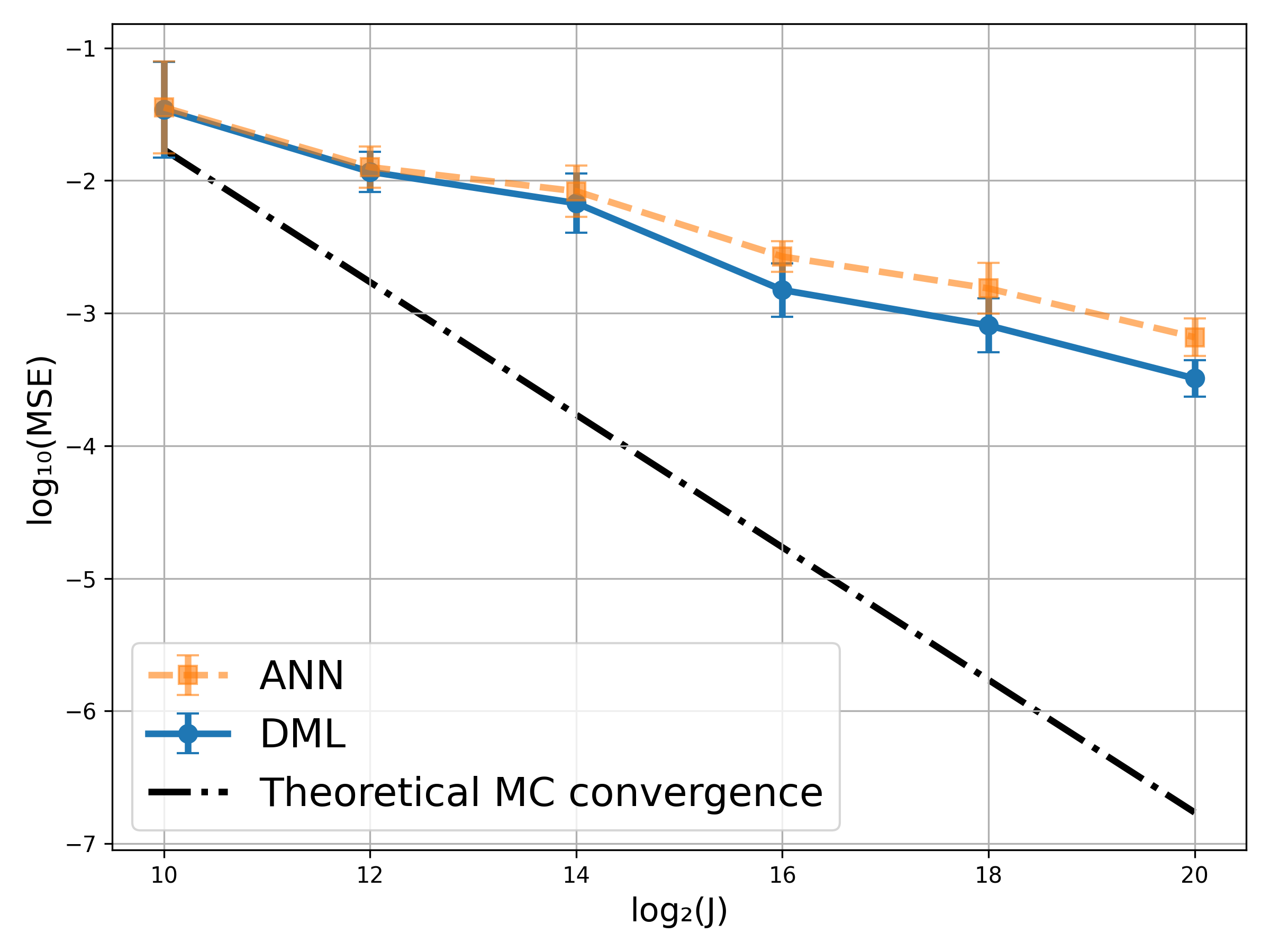}
\caption{MSE vs.\ training set size for the integral arising in the Kou PIDE with \( a = -5 \), \( b = 5 \), \( p \in [0.3,0.7] \), \( \eta_1 \in [3,8] \) and \( \eta_2 \in [1.5,6] \).}
\label{fig:Kou}
\end{figure}

In Figure~\ref{fig:Kou}, we present the convergence results for the integral arising in the Kou PIDE with integration limits \( a = -5 \) and \( b = 5 \), and parameters \( p \in [0.3, 0.7] \), \( \eta_1 \in [3, 8] \) and \( \eta_2 \in [1.5, 6] \). 

\subsubsection{Discussion}

Across all examples in Section~\ref{sec:param_ODEs_PDEs}, several consistent patterns emerge:

\begin{itemize}
    \item \textbf{Accuracy.} The ANN and DML frameworks consistently reproduce analytical or high-precision numerical integrals with high fidelity.  
    \item \textbf{Sample Efficiency.} As illustrated in Figures~\ref{fig:legendre} and~\ref{fig:Kou}, DML consistently yields lower MSEs as the dataset size increases.
    \item \textbf{Scalability.} The advantages of gradient-enhanced training persist as the dimensionality of the parameter vector increases. The method scales smoothly from two parameters in the pendulum example to three in the Kou jump--diffusion model.
\end{itemize}

\section{Conclusions} \label{sec:conclusion}

In this work, a DML-based framework for the numerical approximation of parametric integrals have been investigated. The methodology combines single-realization MC estimation with gradient-enhanced ANN training. By incorporating derivative information of the integrand with respect to model parameters, the proposed DML approach yields differentiable surrogate models that inherit the smoothness and regularity of the underlying mathematical structure.

From a methodological standpoint, it has been demonstrated that single-realization MC labels provide unbiased training data for both function values and parameter differentials. This property enables the construction of data-efficient surrogates capable of approximating entire families of integrals at negligible marginal cost. Theoretical justification is supported by Proposition~\ref{prop:dml_variance_reduction}, which provides unbiasedness and averaged variance reduction via differential learning setting.

Through a series of numerical experiments, the benefits in precision, efficient training set generation and applicability of the proposed DML framework have been clearly evidenced.

All in all, DML-based integration can be regarded as a principled and efficient alternative to classical quadrature methods, MC estimations and standard data-driven surrogates for solving parametric integrals. By simultaneously enhancing accuracy, sample efficiency, and scalability, the approach establishes a rigorous foundation for future developments in data-driven numerical analysis, operator learning, and scientific computing.

\section*{Acknowledgements}

This research has been directly funded via the BERCE internal program, promoted by the University of A Coruña. Both authors acknowledge the funding from the Ministry of Science and Innovation of Spain through the program with reference PID2022-141058OB-I00, and from the Department of Education, Science, Universities, and Vocational Training of the Xunta de Galicia through the program with reference ED451C 2022/047, as well as the support from CITIC, as a centre accredited for excellence within the Galician University System and a member of the CIGUS Network, receiving subsidies from the Department of Education, Science, Universities, and Vocational Training of the Xunta de Galicia. Additionally, it is co-financed by the EU through the FEDER Galicia 2021-27 operational program (ref. ED451G 2023/01). Á. Leitao also acknowledges the funding from the Ministry of Science and Innovation of Spain through the Ramón y Cajal 2022 grant, and from the Department of Education, Science, Universities, and Vocational Training of the Xunta de Galicia through the Excellence program with reference ED431F 2025/032.

\bibliographystyle{plain}
\bibliography{References}

\appendix

\section{Appendices} \label{sec:appendices}

\subsection{Proof of Proposition \ref{prop:dml_variance_reduction}} 
\label{app:proof_prop}

\begin{proof}
The proof proceeds in two steps.

\medskip
\noindent\textbf{Step 1. Unbiasedness.}
By assumptions \textnormal{(A1)}--\textnormal{(A3)} and the dominated convergence theorem, differentiation under the integral sign is justified, and for every 
$\boldsymbol{\theta}'\in\Omega_{\boldsymbol{\theta}}$,
\[
\nabla_{\!\boldsymbol{\theta}} I(\boldsymbol{\theta}')
= \nabla_{\!\boldsymbol{\theta}} \int_{\mathcal X} f(x;\boldsymbol{\theta}')\,\nu(\mathrm d x)
= \int_{\mathcal X} \nabla_{\!\boldsymbol{\theta}} f(x;\boldsymbol{\theta}')\,\nu(\mathrm d x).
\]

Thus,
\begin{equation} \label{eq:expectation}
I(\boldsymbol{\theta}^{(j)}) = \mathbb{E}_\nu\big[\hat{y}^{(j)}\big],
\qquad
\nabla_{\!\boldsymbol{\theta}} I(\boldsymbol{\theta}^{(j)})
 = \mathbb{E}_\nu\big[\hat{g}^{(j)}\big].
\end{equation}

Let $\displaystyle \mathbf{w}_\vartheta\in\operatorname*{argmin}_{\mathbf{w}} \mathcal{L}_\vartheta(\mathbf{w})$
be a minimiser of~\eqref{eq:combined_loss}. The first–order optimality condition yields
\begin{align*}
0 = \frac{\partial}{\partial\mathbf w}\mathcal{L}_\vartheta(\mathbf w_\vartheta)
= \ & 
\frac{2\vartheta}{J}\sum_{j=1}^J \mathbb{E}_\nu\Big[
 \big(\widehat{I}^{(\vartheta)}(\boldsymbol{\theta}^{(j)};\mathbf w_\vartheta) - \hat{y}^{(j)}\big)
 \,\frac{\partial}{\partial \mathbf w}\widehat{I}^{(\vartheta)}(\boldsymbol{\theta}^{(j)};\mathbf w_\vartheta)
\Big]\\
& + \frac{2(1-\vartheta)}{J}\sum_{j=1}^J \mathbb{E}_\nu\Big[
  \big(\nabla_{\!\boldsymbol{\theta}}\widehat{I}^{(\vartheta)}(\boldsymbol{\theta}^{(j)};\mathbf w_\vartheta)
      - \hat{g}^{(j)}\big)
  \cdot \frac{\partial}{\partial \mathbf w}\nabla_{\!\boldsymbol{\theta}}
      \widehat{I}^{(\vartheta)}(\boldsymbol{\theta}^{(j)};\mathbf w_\vartheta)
\Big].
\end{align*}

Rearranging terms, for each $j$,
\small \begin{equation} \label{eq:identity}
\mathbb{E}_\nu\Big[
 \big(\widehat{I}^{(\vartheta)}(\boldsymbol{\theta}^{(j)};\mathbf w_\vartheta)-\hat{y}^{(j)}\big)
 \,\frac{\partial}{\partial \mathbf w}\widehat{I}^{(\vartheta)}(\boldsymbol{\theta}^{(j)};\mathbf{w}_\vartheta)
\Big]
=
-\frac{1-\vartheta}{\vartheta}\,
\mathbb{E}_\nu\Big[
 \big(\nabla_{\!\boldsymbol{\theta}}\widehat{I}^{(\vartheta)}(\boldsymbol{\theta}^{(j)};\mathbf w_\vartheta)
 -\hat{g}^{(j)}\big)
 \,\frac{\partial}{\partial \mathbf{w}}\nabla_{\!\boldsymbol{\theta}}
     \widehat{I}^{(\vartheta)}(\boldsymbol{\theta}^{(j)};\mathbf w_\vartheta)
\Big].
\end{equation} \normalsize

The left-hand identity \eqref{eq:identity} implies the scalar orthogonality relation
\[
\mathbb{E}_\nu\big[
 \widehat{I}^{(\vartheta)}(\boldsymbol{\theta}^{(j)};\mathbf w_\vartheta)
 - \hat{y}^{(j)}
\big] = 0.
\]

Using~\eqref{eq:expectation},
\[
\mathbb{E}_\nu\!\big[\widehat{I}^{(\vartheta)}(\boldsymbol{\theta}^{(j)})\big]
= \mathbb{E}_\nu\!\big[\hat{y}^{(j)}\big]
= I(\boldsymbol{\theta}^{(j)}).
\]

Similarly, the right-hand identity \eqref{eq:identity} yields
\[
\mathbb{E}_\nu\!\big[
 \nabla_{\!\boldsymbol{\theta}}
 \widehat I^{(\vartheta)}(\boldsymbol{\theta}^{(j)})
\big]
=
\mathbb{E}_\nu\!\big[\hat{g}^{(j)}\big]
=
\nabla_{\!\boldsymbol{\theta}} I(\boldsymbol{\theta}^{(j)}),
\]
establishing the unbiasedness statements in (i).

\medskip
\noindent\textbf{Step 2. Variance reduction.}
For each $j$ and $\mathbf w\in\mathbb{R}^{\mathrm m}$. Define the per-sample value-loss
\[
[\mathcal{L}^{(\vartheta)}_{\mathrm{val}}]^{(j)}(\mathbf w)
:= \mathbb{E}_\nu\big[
 (\widehat{I}^{(\vartheta)}(\boldsymbol{\theta}^{(j)};\mathbf w)-\hat{y}^{(j)})^2
\big].
\]

Adding and subtracting $I(\boldsymbol{\theta}^{(j)})$ inside the square and using
$\mathbb{E}_\nu[I(\boldsymbol{\theta}^{(j)})-\hat{y}^{(j)}]=0$
gives
\[
[\mathcal{L}^{(\vartheta)}_{\mathrm{val}}]^{(j)}(\mathbf w)
= \mathbb{E}_\nu\big[
 (\widehat{I}^{(\vartheta)}(\boldsymbol{\theta}^{(j)};\mathbf w)-I(\boldsymbol{\theta}^{(j)}))^2
\big]
+ C_j,
\]
where
\[
C_j = \mathbb{E}_\nu\big[
  (I(\boldsymbol{\theta}^{(j)}) - \hat{y}^{(j)})^2
\big]
\]
is independent of~$\mathbf w$.  

Thus,
\begin{equation} \label{eq:valrisk_decomp}
\mathcal{L}^{(\vartheta)}_{\mathrm{val}}(\mathbf w)
=
\frac{1}{J}\sum_{j=1}^J
\mathbb{E}_\nu\big[
 (\widehat{I}^{(\vartheta)}(\boldsymbol{\theta}^{(j)};\mathbf w)
  - I(\boldsymbol{\theta}^{(j)}))^2
\big]
+ \frac{1}{J}\sum_{j=1}^J C_j.
\end{equation}

By definition, $\mathbf w_\vartheta$ minimizes $\mathcal L_\vartheta$, so evaluating $\mathcal L_\vartheta$ at $\mathbf w_\vartheta$ cannot exceed its value at $\mathbf w$. Thus,
\begin{equation*} 
\mathcal{L}^{(\vartheta)}_{\mathrm{val}}(\mathbf w_\vartheta) \le \mathcal{L}^{(\vartheta)}_{\mathrm{val}}(\mathbf w_{\vartheta}) + \frac{1-\vartheta}{\vartheta} \mathcal{L}^{(\vartheta)}_{\mathrm{diff}}(\mathbf w_{\vartheta}) \le \mathcal{L}^{(\vartheta)}_{\mathrm{val}}(\mathbf w) + \frac{1-\vartheta}{\vartheta} \mathcal{L}^{(\vartheta)}_{\mathrm{diff}}(\mathbf w).
\end{equation*}

Since $\mathbf w_1$ minimizes the pure value-loss $\mathcal L^{(1)}_{\mathrm{val}}$, it therefore follows that
\[
\mathcal{L}^{(\vartheta)}_{\mathrm{val}}(\mathbf w_\vartheta)
\le \mathcal{L}^{(1)}_{\mathrm{val}}(\mathbf w_1).
\]

Using \eqref{eq:valrisk_decomp}, this implies
\begin{equation} \label{eq:inequality}
\frac{1}{J}\sum_{j=1}^J
\mathbb{E}_\nu\!\left[
 \big(\widehat I^{(\vartheta)}(\boldsymbol{\theta}^{(j)})
 - I(\boldsymbol{\theta}^{(j)})\big)^2
\right]
\le
\frac{1}{J}\sum_{j=1}^J
\mathbb{E}_\nu\!\left[
 \big(\widehat I^{(1)}(\boldsymbol{\theta}^{(j)})
 - I(\boldsymbol{\theta}^{(j)})\big)^2
\right].
\end{equation}

From Step~1,
\[
\mathbb{E}_\nu\!\left[\widehat I^{(\vartheta)}(\boldsymbol{\theta}^{(j)})\right]
= I(\boldsymbol{\theta}^{(j)}),
\]
so the inequality \eqref{eq:inequality} gives
\[
\frac{1}{J}\sum_{j=1}^J
\operatorname{Var}_\nu\!\big(\widehat I^{(\vartheta)}(\boldsymbol{\theta}^{(j)})\big)
\le
\frac{1}{J}\sum_{j=1}^J
\operatorname{Var}_\nu\!\big(\widehat I^{(1)}(\boldsymbol{\theta}^{(j)})\big).
\]
establishing the averaged variance-reduction claim (ii).
\end{proof}

\end{document}